\documentclass{article} 
\usepackage{iclr2027_conference,times}

\usepackage{hyperref}
\usepackage{url}

\usepackage{amsmath,amssymb,amsthm}
\usepackage[ruled, noline, scleft]{algorithm2e}
\usepackage{diagbox}
\usepackage{graphicx}
\usepackage{booktabs} 
\usepackage{multirow}
\usepackage{placeins}
\usepackage{wrapfig}
\usepackage{makecell}

\newtheorem{lemma}{Lemma}
\newtheorem{remark}{Remark}
\SetKwComment{Comment}{ // }{}

\newcommand{\vect}[1]{\mathbf{#1}}

\title{Mitigating Multimodal LLMs Hallucinations via Relevance Propagation at Inference Time}

\author{Itai Allouche, Joseph Keshet\\
Faculty of Electrical and Computer Engineering,\\
Technion, Haifa, Israel\\
\texttt{itai208@campus.technion.ac.il}
}

\iclrfinalcopy 
\begin{document}

\maketitle
\lhead{} 
\begin{abstract}
Multimodal large language models (MLLMs) achieve strong performance on vision- and audio-language tasks, yet can generate responses that conflict with the given visual or auditory inputs, a problem known as multimodal hallucinations. Prior work suggests that this occurs when models rely more on textual cues and learned language patterns than on evidence from the perceptual input. 
To obtain a more direct account of this imbalance, we apply Layer-wise Relevance Propagation (LRP), which attributes predictions to individual input tokens, and use the resulting relevance scores to analyze and mitigate hallucinations. First, we examine whether this imbalance leads to multimodal hallucinations. We find that hallucinations often arise when the model relies less on perceptual inputs, and that changing this reliance affects its predictions. We further leverage LRP and propose a training-free framework that shifts relevance toward perceptual tokens by optimizing key-value representations during decoding, without modifying model parameters or requiring training data. We call this method Learning Inference-time Modality Enhancement (LIME). Despite using no spatial or temporal supervision, LIME concentrates relevance on query-relevant regions. We evaluate LIME across multiple multimodal benchmarks in both vision and audio domains, demonstrating consistent reductions in hallucinations and enhanced grounding while preserving generation quality.
\end{abstract}

\section{Introduction}
\label{sec:introduction}

Multimodal Large Language Models (MLLMs) have recently demonstrated remarkable capabilities in integrating language with perceptual modalities such as images~\citep{liu2023visual, Qwen-VL, Qwen2.5-VL} and audio~\citep{chu2024qwen2, tang2024salmonn}, enabling strong performance on tasks including visual question answering, audio reasoning, and multimodal dialogue. By coupling powerful large language models (LLMs) with perceptual-specific encoders, these systems are able to generate fluent and context-aware responses grounded in diverse sensory inputs. Despite these advances, MLLMs remain prone to hallucinations, producing outputs that are inconsistent with or unsupported by the provided multimodal evidence~\citep{liu2024survey, huang2024visual, zheng2025reefknot}.

A growing body of empirical evidence suggests that multimodal hallucinations are related to an imbalance in how models utilize different input modalities during inference~\citep{liu2024paying, zoulook, hsu2025reducing}. In many settings, text tokens dominate generation, while visual or auditory inputs are under-utilized. \citet{asadi2026mirage} indicated that the apparent visual reasoning capabilities of multimodal models may, in part, arise from strong textual priors, with models producing plausible visual descriptions even without access to the corresponding inputs. However, this imbalance is often inferred through indirect signals, leaving unclear whether reduced reliance on perceptual information actually contributes to hallucination or merely accompanies it.

\begin{figure}[t]
  \centering
  \includegraphics[width=\linewidth]{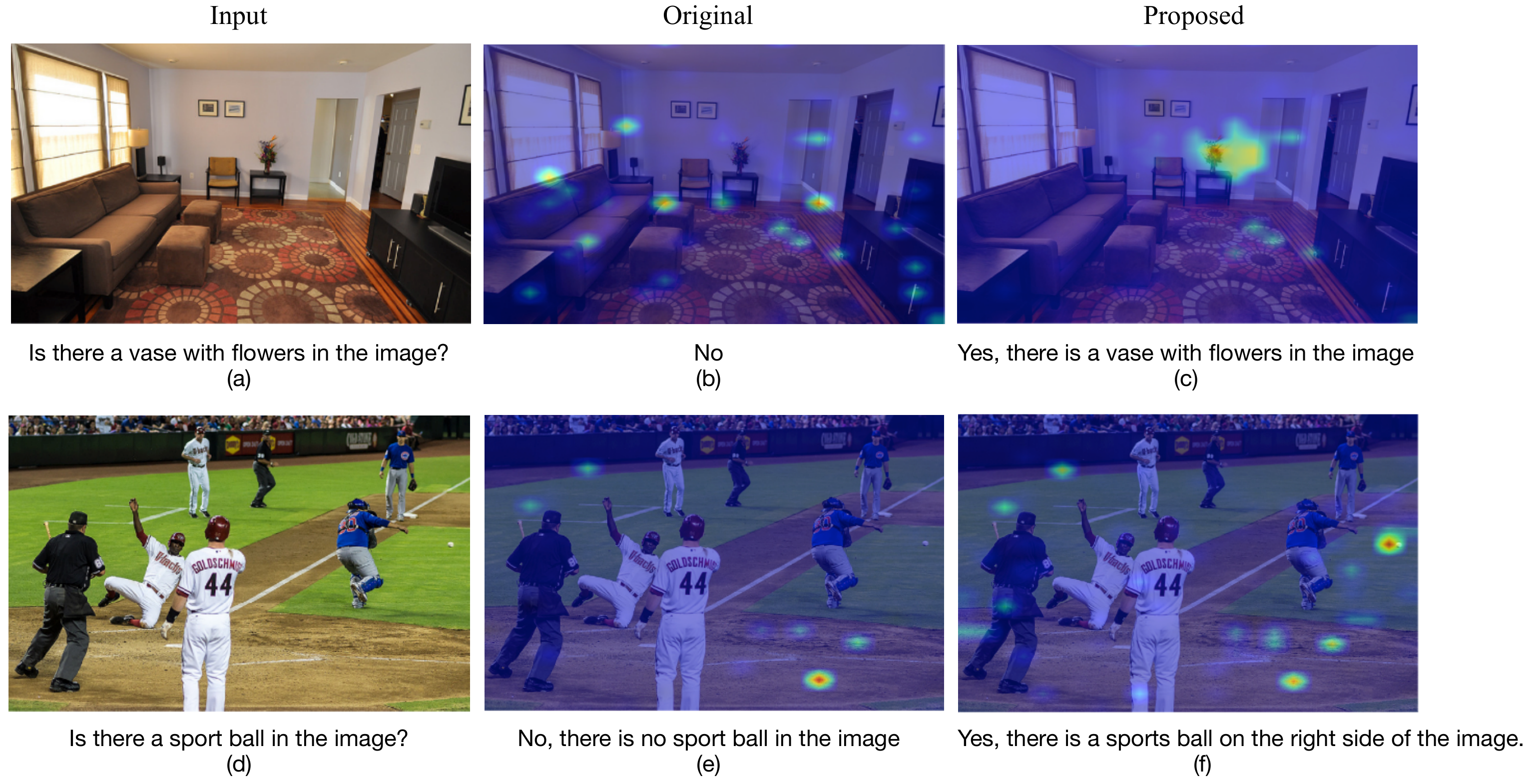}
    \caption{Examples of multimodal hallucinations and their mitigation with LIME. Panels (a, d) show the inputs and questions. Under standard decoding (b, e), the model fails to groundits answers in the visual evidence, while with LIME (c, f) it localizes the relevant regionsand predicts correctly.}
  \label{fig:visionExamples}
\end{figure}

To investigate this question, we employ Layer-wise Relevance Propagation (LRP)~\citep{bach2015pixel}, an interpretability method that decomposes model predictions into the additive contributions of individual input tokens. LRP provides fine-grained \emph{relevance} scores that quantify how strongly each token contributes to a given output. Our analysis shows that LRP uncovers a systematic imbalance in MLLMs: textual tokens dominate the relevance assigned to the input, and what perceptual tokens receive is spread diffusely rather than concentrated on the query-relevant evidence, as illustrated in Figure~\ref{fig:visionExamples}(b) and (e). We further observe that both patterns tend to be most pronounced on samples the model gets wrong, and that intervening on this distribution affects whether the model hallucinates.

Motivated by these findings, this paper focused on mitigating this imbalance at inference time, without additional training. Prior training-free methods pursue this goal indirectly, through raw attention weights or output shifts under perturbed inputs, without measuring whether perceptual reliance actually changes~\citep{hsu2025reducing, leng2024mitigating, liu2024paying, wang2024mitigating}. We instead use LRP relevance itself as the guiding signal, as it directly quantifies each token's contribution rather than relying on proxies such as attention weights. To the best of our knowledge, this is the first work to use LRP, traditionally a post-hoc explanation method, as an inference-time guidance signal in MLLMs.

Specifically, we turn the model's relevance signal into an optimization objective that shifts relevance toward perceptual tokens by modifying key-value representations during decoding. The optimization is performed locally at each decoding step while the underlying model parameters remain fixed. We term this method \emph{Learning Inference-time Modality Enhancement (LIME)}. Despite using no spatial or temporal supervision, LIME increases perceptual utilization and promotes more focused relevance on query-relevant regions. Further analysis shows that its annotation-free objective approaches the localization performance of a region-supervised variant. We evaluate LIME across multiple MLLMs and tasks in both vision and audio domains, demonstrating consistent reductions in hallucination and improved localization.

In summary, our contributions are the following: (i) We provide an LRP-based analysis showing that hallucinations are associated with reduced perceptual relevance, and demonstrate through targeted interventions that altering this relevance systematically changes prediction behavior; (ii) We propose a self-supervised, relevance-guided inference-time framework that mitigates hallucinations by increasing the contribution of perceptual tokens, while concentrating relevance on query-relevant regions without spatial or temporal supervision; (iii) We demonstrate the effectiveness and generality of our approach across multiple multimodal models and benchmarks. Implementation details and source code are publicly available.\footnote{https://github.com/ItaiAllouche/lime} 

\begin{figure}[t]
  \centering
  \includegraphics[width=\linewidth]{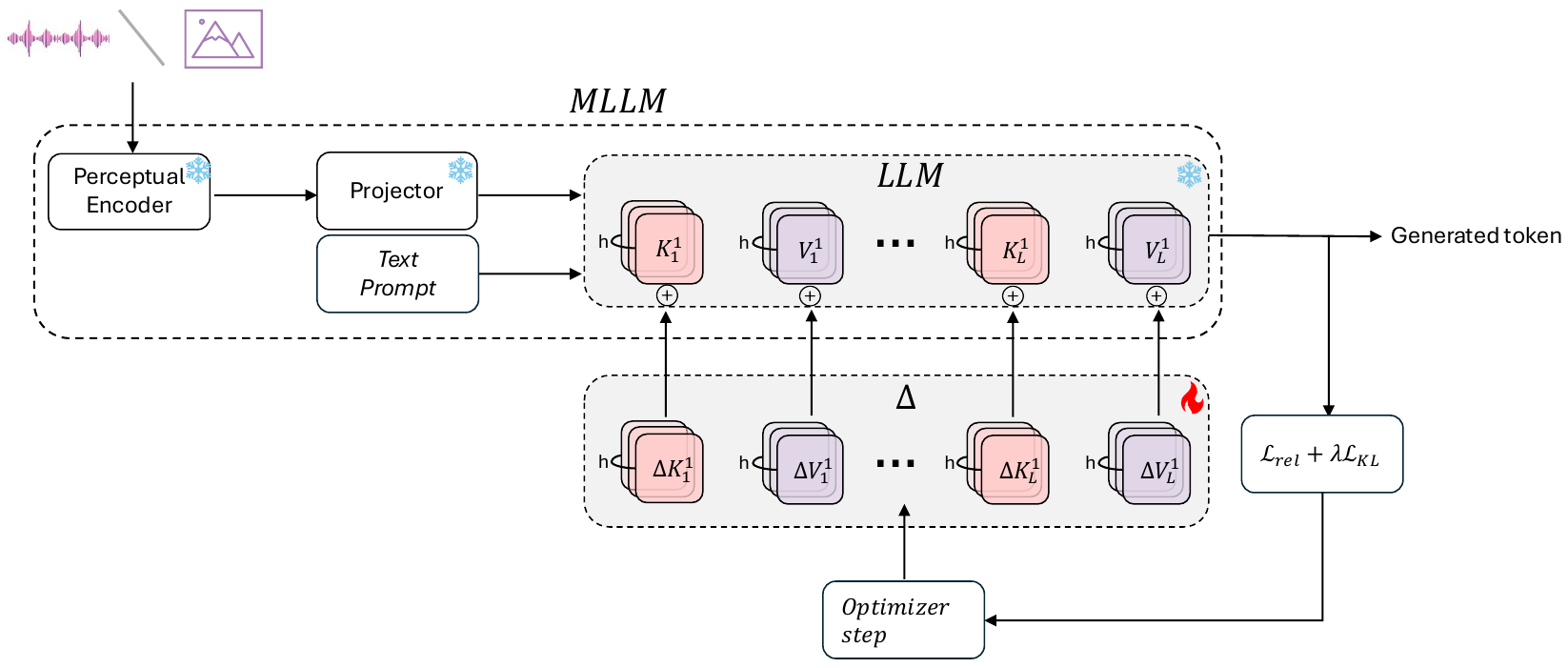}
  \caption{LIME iteratively increases the modality relevance ($\mathcal{L}_{rel}$) while remaining close
to the original model distribution ($\mathcal{L}_{KL}$). The base MLLM is frozen; learning
occurs only through optimizable $\Delta$. Snowflake and flame symbols
denote frozen and optimizable variables, respectively.}
  \label{fig:methodOverview}
\end{figure}

\section{Related Work}
\label{sec:relatedWork}

Hallucination in MLLMs refers to the generation of content that is inconsistent with the provided perceptual inputs, such as describing nonexistent objects in images or hallucinating events and entities in audio. This phenomenon has been widely observed across vision- and audio-language models~\citep{liu2024paying, zoulook, zheng2025reefknot, kuan2025can}. Existing approaches for mitigating multimodal hallucinations can be broadly categorized into training-based and training-free methods. Training-based approaches aim to improve perceptual utilization by updating model parameters through supervised fine-tuning~\citep{gunjal2024detecting}, reinforcement learning~\citep{yu2024rlhf}, or auxiliary modules that revise hallucinated outputs~\citep{zhou2024analyzing}. In contrast, training-free methods intervene directly during inference without modifying model weights. A prominent line of training-free work focuses on attention-based interventions. For example, OPERA~\citep{huang2024opera} penalizes dominant tokens to encourage attention toward perceptual inputs. Another class of methods is based on contrastive decoding (CD)~\citep{li2023contrastive}, which suppresses tokens that are insensitive to perceptual information by comparing outputs under perturbed or modality-removed inputs. Representative approaches include PAI~\citep{liu2024paying}, Visual Contrastive Decoding (VCD)~\citep{leng2024mitigating}, Instruction Contrastive Decoding (ICD)~\citep{wang2024mitigating}, and Audio-Aware Decoding (AAD)~\citep{hsu2025reducing}. More recent methods, such as MemVR~\citep{zoulook} and V-ITI~\citep{sun2025v}, modify hidden representations during inference to improve grounding. Despite notable progress, existing training-free methods often rely on heuristic intervention rules or indirect signals, and do not explicitly quantify the contribution of different modalities to the final prediction. This leaves the relationship between modality contribution and hallucination behavior less directly examined.

Concurrently,~\citet{Elisha_2026_CVPR} proposed Concept-Guided Fine-Tuning (CFT), which uses LRP-based relevance maps to align model explanations with semantic concepts and reduce spurious correlations. Unlike CFT, which uses relevance to fine-tune model parameters, our method uses relevance both to characterize modality utilization and to directly guide inference-time optimization, without updating model parameters or requiring labeled training data.

\section{Method}
\label{sec:method}

We introduce an inference-time framework that uses LRP relevance to guide decoding by increasing the contribution of perceptual tokens. At each decoding step, the method iteratively modifies the model's key and value representations based on the relevance signal, while keeping the underlying model parameters fixed. We first introduce the notation and model setup, then formulate token-level relevance, and finally describe the inference-time optimization. An overview of the framework is provided in Figure~\ref{fig:methodOverview}.

\subsection{Notation and Setup}
\label{sec:notationAndSetup}

Consider an MLLM composed of a perceptual encoder, a projection module, and an LLM backbone, as shown in Figure~\ref{fig:methodOverview}. Given a multimodal input, the perceptual encoder first transforms the raw perceptual signal (e.g., image or audio) into latent representations, which are then mapped through a projection layer into a sequence of perceptual tokens. Hence, the perceptual tokens are represented in the same space as the embedding of the textual tokens.

Let $\vect{x}_{i}$ denote the embedding vector of the $i$-th input token, and let $\vect{X}=(\vect{x}_1, \ldots, \vect{x}_N)$ denote the sequence of input token embeddings. We partition the input token indices into two disjoint sets: $\mathcal{P}$ for perceptual token and $\mathcal{T}$ for textual tokens. Let $\vect{y}_{<t}=(y_1, \dots ,y_{t-1})$ denote the sequence of generated tokens prior to decoding step $t$. At step $t$, the model predicts the next-token distribution $p_{\theta}(y_t \mid \vect{X}, \vect{y}_{<t})$ with corresponding logit vector $\vect{z}_t$. We define $z_t^* = \vect{z}_t[y_t]$ as the scalar logit associated with the generated token $y_t$.

\subsection{Relevance and Problem Formulation}
\label{sec:relevanceProblemFormulation}

We define a relevance function $\Phi(z_t^*)$ that assigns a scalar relevance score to each token in $\vect{X}$, with respect to the output logit $z_t^*$. For brevity, we omit the explicit dependence on $z_t^*$ in the notation of $\Phi$.  Let $\Phi_i$ denote the relevance assigned to the $i$-th token in $\vect{X}$.
 
Given the output logit $z_t^*$, relevance is propagated backward from the output to the input tokens. Following the basic LRP formulation~\citep{bach2015pixel}, the relevance score $\Phi_{j}^{\ell}$ of unit $j$ at layer $\ell$ is computed recursively from the relevance $\Phi_{i}^{\ell + 1}$ of units $i$ at the next layer $\ell +1$ 
\[
    \Phi_{j}^{\ell}=
    \sum_{i}\frac{a_{j}^{\ell} W_{ji}^{\ell}}{\sum_{k}a_{k}^{\ell} W_{ki}^{\ell}} \Phi_{i}^{\ell +1},
\]
where $\vect{a^{\ell}}$ denotes the input to the $\ell$-layer, and $\vect{W}^{\ell}$ denotes the weights associated with layer $\ell$. The denominator ensures proper normalization of relevance. Relevance at lower layers is obtained by recursively applying this rule. The full transformer-specific relevance formulation, including the rules used for attention and other architectural components, is provided in Appendix~\ref{app:layerWiseRelevancePropogation}.

In our setting, relevance propagation is performed through the LLM backbone to obtain token-level relevance scores over $\vect{X}$. We define the total relevance assigned to perceptual and textual tokens as:
\begin{equation}
    \label{eq:totalRelevance}    
    \Phi_{\mathcal{P}} =
    \sum_{i \in \mathcal{P}} \Phi_{i} \quad  
    \Phi_{\mathcal{T}} =
    \sum_{j \in \mathcal{T}} \Phi_{j}.
\end{equation}
These quantities capture the overall contribution of perceptual and textual tokens to the prediction. We use the aggregated quantities $\Phi_{\mathcal{P}}$ and $\Phi_{\mathcal{T}}$ to measure the relative contribution of perceptual and textual tokens. Our objective is therefore to increase the contribution of the \emph{relevant} perceptual tokens during decoding by shifting relevance toward $\mathcal{P}$, while keeping the model close to its original behavior. We further analyze how this relevance redistribution relates to hallucination behavior in Section~\ref{sec:modalityUtilizationAnalysis}.

\subsection{Inference-Time Optimization}
\label{sec:inferenceTimeOptimization}

We define $\Delta =\{\Delta \vect{K},\Delta \vect{V}\}$ as additive updates to the key and value tensors across all transformer layers during inference. The updates have the same dimensionality as the corresponding key and value tensors and are applied across the full sequence rather than to specific token positions. These updates are computed independently at each decoding step and reset afterward.

To motivate our unsupervised relevance objective, consider first the \emph{supervised} setting in which the query-relevant perceptual tokens are known. Let $\mathcal{G} \subseteq \mathcal{P}$ denote this subset, corresponding, for example, to image patches within the queried object's region or audio tokens within the relevant temporal segment. This set be considered as the \emph{true label} our model should enhance. 
We formulate the supervised relevance loss as a cross-entropy objective:
\begin{equation}
    \label{eq:supervisedrelevanceLoss}
    \mathcal{L}^{sup}_{rel}(\Delta)=
    -\frac{1}{|\mathcal{G}|}
    \sum_{i\in \mathcal{P}} 
    {1}_{\{i \in \mathcal{G}\}}
    \log \frac{\exp(\Phi_{i, \Delta} / \tau)}
    {\sum_{k\in \mathcal{P} \cup \mathcal{T}}\exp(\Phi_{k, \Delta} / \tau)},
\end{equation}
where  $\Phi_{i, \Delta}$ denotes the relevance of the $i$-th token under the key and value updates, $\tau > 0$ is a temperature parameter, and ${1}_{\{\cdot\}}$ is an indicator function. Minimizing this objective increases the relative relevance assigned to the query-relevant perceptual tokens $\mathcal{G}$.

However, identifying $\mathcal{G}$ requires spatial or temporal annotations, which are unavailable in the inference setting considered here. We therefore relax the supervised objective by replacing $\mathcal{G}$ with the full set of perceptual tokens $\mathcal{P}$, treating each perceptual token as potential relevant evidence. This yields our annotation-free relevance objective
\begin{equation}
    \label{eq:relevanceLoss}
    \mathcal{L}_{rel}(\Delta)=
    -\frac{1}{|\mathcal{P}|}
    \sum_{i\in \mathcal{P}} \log \frac
    {\exp(\Phi_{i, \Delta} / \tau)}
    {\sum_{k\in \mathcal{P} \cup \mathcal{T}}\exp(\Phi_{k, \Delta} / \tau)}.
\end{equation}
The resulting objective is self-supervised: its guidance is derived entirely from the model's own relevance decomposition and requires no knowledge of which perceptual tokens are relevant to the query. Despite this coarser supervision, we find that the induced relevance becomes increasingly concentrated on query-relevant regions, approaching the localization achieved by the supervised formulation, as analyzed in Section~\ref{sec:modalityUtilizationAnalysis}. To explain why this localization emerges, we provide a local analysis in Appendix~\ref{app:mechanism}, showing how the frozen model’s sensitivity structure filters the broadly distributed objective, resulting in relevance changes that tend to align with query-relevant evidence.

\begin{figure}[t]
  \centering
  \includegraphics[width=\linewidth]{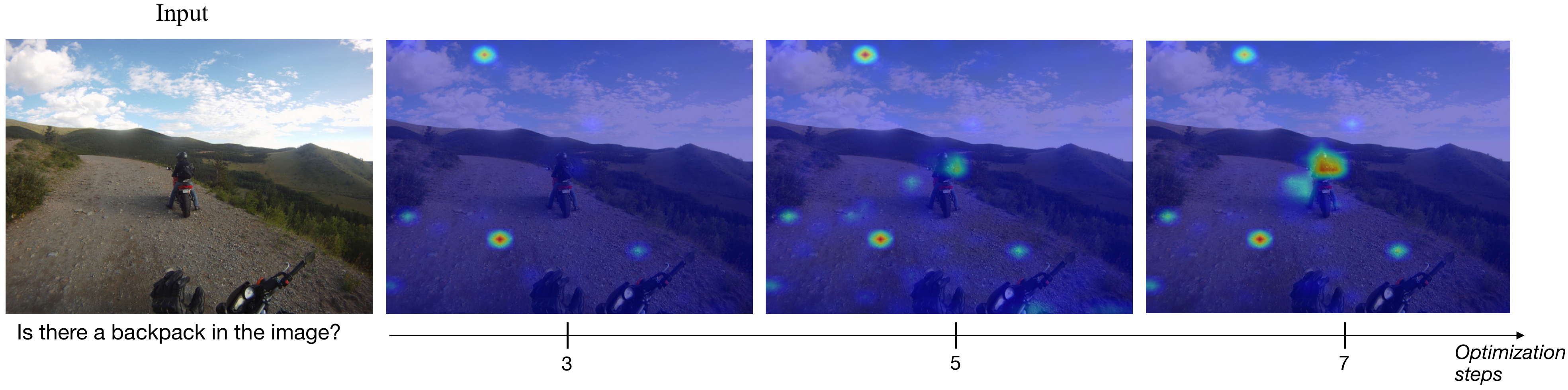}
  \caption{Evolution of visual relevance during inference-time.
    Relevance over the image is shown across LIME optimization steps. Relevance is initially diffuse and becomes progressively concentrated on relevant regions, indicating improved grounding.}
  \label{fig:evolutionVIsualRelevance}
\end{figure}

To constrain deviations from the pretrained model, we use a KL-divergence regularizer between the next-token distributions produced with and without the key-value updates, denoted by $p_{\theta,\Delta}(y_t \mid \vect{X},\vect{y}_{<t})$ and $p_{\theta}(y_t \mid \vect{X}, \vect{y}_{<t})$, respectively. We define
\begin{equation}
    \label{eq:klLoss}
    \mathcal{L}_{KL}(\Delta)=D_{KL}(p_{\theta, \Delta}(y_t \mid \vect{X}, \vect{y}_{<t})\text{ }||\text{ }p_{\theta}(y_t \mid \vect{X}, \vect{y}_{<t})).
\end{equation}
This term constrains modifications to the key and value representations to preserve the pretrained model's linguistic prior and decoding stability.

Combining both loss terms (Eq.~\eqref{eq:relevanceLoss} and Eq.~\eqref{eq:klLoss}), the updates $\Delta$ of $\vect{K}\vect{V}$ are optimized at each decoding step by minimizing the following composite loss:
\begin{equation}
    \label{eq:objective}
    \arg \underset{\Delta}{\mathrm{min}}\text{ }
    \mathcal{L}_{rel}(\Delta) + \lambda \mathcal{L}_{KL}(\Delta),
\end{equation}
where $\lambda > 0$ controls the trade-off between promoting multimodal relevance and constraining deviations from the reference model. Minimizing this objective aims to promote key and value representations that amplify the explanatory responsibility of perceptual tokens while maintaining linguistic coherence and decoding stability. The key and value perturbations $\Delta$ are initialized to zero and updated using a small number of gradient-based optimization steps. Since the relevance objective depends on LRP relevance, optimizing $\Delta$ requires second-order derivatives through the relevance computation. Importantly, these updates are discarded after each step, ensuring that control remains local and does not accumulate across time.

\begin{table}[t]
    \caption{POPE results with LLaVA-1.5-7B on MSCOCO. Accuracy (Acc) and F1 are reported for the random, popular, and adversarial splits and their average. Best and second-best results are in \textbf{bold} and \underline{underlined}, respectively.}
    \label{tab:llavaPope}
    \centering
    \resizebox{\linewidth}{!}{
        \begin{tabular}{llcccccccc}
        \toprule
        \multirow{2}{*}{\textbf{Dataset}} & \multirow{2}{*}{\textbf{Methods}} 
        & \multicolumn{2}{c}{\textbf{Random (\%)} $\uparrow$} 
        & \multicolumn{2}{c}{\textbf{Popular (\%)} $\uparrow$} 
        & \multicolumn{2}{c}{\textbf{Adversarial (\%)} $\uparrow$}
        & \multicolumn{2}{c}{\textbf{Average (\%)} $\uparrow$} \\
        \cmidrule(lr){3-4}
        \cmidrule(lr){5-6}
        \cmidrule(lr){7-8}
        \cmidrule(lr){9-10}
        & & Acc & F1 & Acc & F1 & Acc & F1 & Acc & F1 \\
        \midrule
        \multirow{7}{*}{MSCOCO}
        & LLaVA-1.5-7B & 83.49 & 82.28 & 79.98 & 79.34 & 76.03 & 76.26 & 79.83 & 79.29 \\
        & + OPERA~\citep{huang2024opera} & 87.53 & 86.45 & 84.21 & 83.5 & 80.88 & 80.69 & 84.21 & 83.55 \\
        & + VCD~\citep{leng2024mitigating} & 86.84 & 86.83 & 82.65 & 83.37 & 77.31 & 79.28 & 82.27 & 83.16 \\
        & + ICD~\citep{wang2024mitigating} & 84.87 & 83.27 & 82.93 & 81.45 & 81.07 & 79.96 & 82.96 & 81.56 \\
        & + MemVR~\citep{zoulook} & 88.5 & 87.34 & \underline{87.1} & \underline{86.01} & 85.2 & 84.28 & 86.93 & \underline{85.88} \\
        & + V-ITI~\citep{sun2025v} & \underline{89.74} & \underline{87.72} & 84.96 & 84.77 & \textbf{86.31} & 82.44 & \underline{87} & 84.98 \\
        & + LIME (ours) & \textbf{90.27} & \textbf{89.75} & \textbf{87.91} & \textbf{87.85} & \underline{85.51} & \textbf{84.52} & \textbf{87.89} & \textbf{87.37} \\
        \bottomrule    
        \end{tabular}
    }
\end{table}

\section{Experiments}
\label{sec:experiments}

We next present the empirical evaluation of our approach. We first report results on vision and audio benchmarks, then examine the relationship between modality relevance and hallucination behavior, and finally provide an ablation study. Details on datasets, evaluation protocols, and metrics are deferred to Appendix~\ref{app:evaluationBenchmarks}. Relevance scores are computed using Attention-Aware Layer-wise Relevance Propagation (AttnLRP)~\citep{pmlr-v235-achtibat24a}, which adapts LRP to transformer architectures through attention-specific propagation rules. Implementation details and hyperparameter settings are given in Appendix~\ref{app:baselinesAndImplementationDetails}. All experiments were conducted on 8 NVIDIA A100 GPUs.

\subsection{Vision Benchmarks}
\label{sec:visionBenchmarks}

The five MLLMs we used for vision evaluation were LLaVA-1.5-7B~\citep{liu2023visual}, Qwen-VL-Chat~\citep{Qwen-VL}, Qwen2.5-VL-7B-Instruct~\citep{Qwen2.5-VL}, Qwen2.5-Omni-7B~\citep{Qwen2.5-Omni}, and MiniCPM-o 4.5~\citep{cui2026minicpm}. LLaVA-1.5-7B and the Qwen-based models employ CLIP~\citep{radford2021learning} as visual encoder, paired with LLaMA~\citep{touvron2023llama} and Qwen~\citep{bai2023qwen} LLM backbones respectively, while MiniCPM-o 4.5 uses LLaVA-UHD~\citep{guo2024llava} with a Qwen3-8B backbone. We used their pretrained, instruction-tuned variants without additional fine-tuning, adhering to a strictly zero-shot setting. We compared LIME against the training-free baselines OPERA~\citep{huang2024opera}, VCD~\citep{leng2024mitigating}, ICD~\citep{wang2024mitigating}, MemVR~\citep{zoulook}, and V-ITI~\citep{sun2025v}, reporting comparisons where reproducible, since some are not applicable across all architectures.

The first benchmark we considered was POPE~\citep{li2023evaluating}, which frames hallucination detection as a binary yes/no question-answering task constructed from the MSCOCO~\citep{lin2014microsoft} and A-OKVQA~\citep{schwenk2022okvqa} datasets, using three splits: random, popular, and adversarial (see 
Appendix~\ref{app:evaluationBenchmarks} for more details). Table~\ref{tab:llavaPope} reports accuracy and F1 for LLaVA-1.5-7B across all baselines on MSCOCO images for each split. Our proposed approached, LIME, outperforms all baselines on most metrics, achieving the highest average accuracy and F1, with the largest improvements over the base model on the adversarial split, indicating increased robustness to challenging negative samples. A similar trend holds for other models and for A-OKVQA (Appendix~\ref{app:additionalExperimentsAndResults}).

The second vision benchmark we considered was CHAIR~\citep{rohrbach-etal-2018-object}, which quantifies hallucinations by evaluating object mentions in generated captions on MSCOCO. Performance is measured using CHAIR$_S$, the percentage of sentences containing at least one hallucinated object, and CHAIR$_I$, the fraction of hallucinated mentions among all generated object instances (lower is better for both). As reported in Table~\ref{tab:visionCahir}, our method reduces hallucination rates across all evaluated models, indicating improved utilization of visual tokens. While some methods attain higher recall, they often do so at the expense of increased hallucinations; in contrast, our approach achieves a more favorable trade-off between faithfulness and coverage.

Finally, we evaluate LIME on HallusionBench~\citep{guan2024hallusionbench}, extending our analysis beyond object presence and captioning to diverse image-context reasoning scenarios. As reported in Table~\ref{tab:hallusionBench}, LIME improves both LLaVA-1.5 and Qwen2.5-VL-7B-Instruct across all reported metrics. Overall, these results demonstrate that our method effectively mitigates visual hallucinations while maintaining strong performance across models and datasets.

\begin{table}[t]
    \caption{CHAIR evaluation results. CHAIR$_S$, CHAIR$_I$, their average, and Recall are reported. Best and second-best results are in \textbf{bold} and \underline{underlined}, respectively.}
    \label{tab:visionCahir}
    \centering
    \begin{tabular}{lcccc}
    \toprule
    \textbf{Methods} & \textbf{CHAIR$_S \downarrow$} & \textbf{CHAIR$_I \downarrow$} & \textbf{Average} $\downarrow$ &  \textbf{Recall} $\uparrow$ \\
    \hline
    LLaVA-1.5-7B & 52 & 15.8 & 32.7 & 75.2 \\
    + OPERA & 47.8 & 14.6 & 31.8 & 77.3 \\
    + VCD & 48.6 & 14.9 & 31.2 & 76.8 \\
    + ICD & 56.2 & 16.3 & 36.3 & 16.3 \\
    + MemVR & 46.6 & \textbf{13} & \underline{29.8} & \textbf{80.8} \\
    + V-ITI & \underline{46.1} & \underline{13.5} & \underline{29.8} & \underline{80.4} \\
    + LIME (ours) & \textbf{42.7} & \textbf{13} & \textbf{27.85} & 72 \\
    \hline
    Qwen-VL-Chat & 46 & 12.5 & 29.3 & 64.3 \\
    + VCD & 46.8 & \underline{12.3} & 29.6 & \underline{67.9} \\
    + ICD & 45 & 14.3 & 29.7 & 47.6 \\    
    + V-ITI & \textbf{44.2} & 12.5 & \underline{28.4} & 66.4 \\
    + LIME (ours) & \underline{44.5} & \textbf{12} & \textbf{28.25} & \textbf{68.7} \\
    \hline
    Qwen2.5-VL-7B-Instruct & 25.6 & 9.1 & 17.35 & 55.1 \\
    + LIME (ours) & \textbf{21.2} & \textbf{8.2} & \textbf{14.7} & \textbf{56.5} \\ 
    \bottomrule
    \end{tabular}
\end{table}

\subsection{Audio Benchmarks}
\label{sec:audioBenchmarks}

The three MLLMs we used for audio evaluation were
SALMONN-7B~\citep{tang2024salmonn}, Qwen2-Audio-7B-Instruct~\citep{chu2024qwen2}, and Qwen2.5-Omni-7B. SALMONN-7B employs a Whisper~\citep{radford2023robust} encoder large v2 paired with Vicuna~\citep{vicuna2023} LLM backbone. Both Qwen2-Audio-7B-Instruct and Qwen2.5-Omni-7B also use Whisper encoder large v3 paired with Qwen LLM backbone. We used the default configurations from the original papers for all methods. Our proposed model is compared against AAD~\citep{hsu2025reducing}, the only hallucination mitigation training-free method we found. The results reported for this model are based on our own implementation, as the original authors have not released their code.

The first audio benchmark, Audio Hallucination QA (AHQA)~\citep{kuan2024understanding}, formulates hallucination detection as a binary question-answering task: given an audio clip, models are asked yes/no questions of the form ``Is there a sound of [object] in the audio?''. The second audio benchmark was AIR-Bench~\citep{yang2024air}, which evaluates generative audio understanding across three domains: speech, sound, and music. In this setting, questions are presented in a multiple-choice format, but models must generate answers freely, making the evaluation fully generative.

Results summarized in Table~\ref{tab:audioEvals} and Appendix~\ref{app:additionalExperimentsAndResults}. On AHQA, our method improves both accuracy and F1 score across all sampling strategies. For Qwen2-Audio-7B-Instruct, we observe substantial gains over prior methods, particularly in F1 score. Similar improvements are observed for SALMONN-7B and Qwen2.5-Omni-7B. On AIR-Bench, our method improves performance across all domains, with consistent gains in speech and sound understanding and competitive performance in music tasks. These results suggest that LIME not only reduces hallucinations but also enhances general multimodal reasoning in audio settings.

\begin{table}[t]
    \caption{Accuracy and F1 scores for the Audio Hallucination QA and Air-Bench benchmarks.}
    \label{tab:audioEvals}
    \centering
    \resizebox{\columnwidth}{!}{%
    \begin{tabular}{lccccccccc}
        \toprule
        \diagbox{\textbf{Methods}}{\textbf{Dataset}}
        & \multicolumn{6}{c}{\textbf{AHQA} \textbf{(\%) $\uparrow$}}
        & \multicolumn{3}{c}{\textbf{Air-Bench} \textbf{(\%) $\uparrow$}} \\
        \cmidrule(lr){2-7}\cmidrule(lr){8-10}
        
        & \multicolumn{2}{c}{Random}
        & \multicolumn{2}{c}{Popular}
        & \multicolumn{2}{c}{Adversarial}
        & \multirow{2}{*}{Speech}
        & \multirow{2}{*}{Sound}
        & \multirow{2}{*}{Music} \\
        \cmidrule(lr){2-3}\cmidrule(lr){4-5}\cmidrule(lr){6-7}
        
        & Acc & F1
        & Acc & F1
        & Acc & F1
        &  &  & \\
        \midrule
        
        SALMONN-7B & 53.91 & 23.37 & 49.32 & 18.27 & 50.31 & 20.01 & 37.51 & 33.58 & 31.05 \\
        + AAD~\citep{hsu2025reducing} & \textbf{57.22} & 36.74  & 48.71 & 18.78 & 48.04 & 17.42 & 42.62 & 34.56 & 30.38 \\
        + LIME (ours) & 56.88 & \textbf{36.85}  & \textbf{53.12} & \textbf{25.76} & \textbf{54.35} & \textbf{26} & \textbf{45.2} & \textbf{36.9} & \textbf{31.95} \\
        \midrule
        Qwen2-Audio-7B-Instruct & 56.19 & 26.78  & 51.34 & 20.5 & 50.13 & 20.24 & 57.56 & 60.86 & 55.89 \\
        + AAD~\citep{hsu2025reducing} & 59.5 & 31.62 & 51.31 & 12.09 & 51.29 & 11.43 & 60 & 61.9 & \textbf{57.43} \\
        + LIME (ours) & \textbf{63.36} & \textbf{50.27} & \textbf{57.53} & \textbf{46.43} & \textbf{53.1} & \textbf{37.08} & \textbf{66.1} & \textbf{66.41} & 56.25 \\      
        \bottomrule
    \end{tabular}
    }
\end{table}

\begin{table}[t]
\centering
\small
\setlength{\tabcolsep}{4pt}
\caption{Perceptual utilization and relevance localization across models and configurations.}
\label{tab:modalityUtilization}
\begin{tabular}{lccccccccc}
\toprule
& \multicolumn{2}{c}{\textbf{Perceptual Utilization $\uparrow$}}
& \multicolumn{7}{c}{\textbf{Relevance Localization $\uparrow$}} \\
\cmidrule(lr){2-3}\cmidrule(lr){4-10}
& & & \textbf{Baseline} & \multicolumn{3}{c}{\textbf{LIME}}
& \multicolumn{3}{c}{\textbf{LIME-Sup}} \\
\cmidrule(lr){4-4}\cmidrule(lr){5-7}\cmidrule(lr){8-10}
\textbf{Model} & \textbf{Baseline} & \textbf{LIME} & --
& \textbf{3} & \textbf{5} & \textbf{7} & \textbf{3} & \textbf{5} & \textbf{7} \\
\midrule
LLaVA-1.5-7B   & 0.10 & 0.17 & 0.27 & 0.29 & 0.315 & 0.351 & 0.32 & 0.361 & 0.37 \\
Qwen2-Audio-7B & 0.34 & 0.42 & 0.31 & 0.4 & 0.5 & 0.57 & 0.491 & 0.56 & 0.585 \\
\bottomrule
\end{tabular}
\end{table}

\subsection{Modality Utilization Analysis}
\label{sec:modalityUtilizationAnalysis}

To better understand how modality utilization changes under LIME, we conduct a relevance-based analysis using LRP. Since LRP conserves relevance~\citep{bach2015pixel}, LIME cannot change how much relevance exists, only how it is distributed among tokens. Here, we ask two questions about this distribution. First, whether relevance shifts from textual to perceptual tokens relative to standard decoding. Second, how the increased perceptual relevance is distributed within the perceptual input: since Eq.~\eqref{eq:relevanceLoss} treats all perceptual tokens symmetrically and uses no spatial or temporal supervision, the shift could in principle be spread uniformly rather than concentrated on the evidence relevant to the query.  Separately, Appendix~\ref{app:interventional} examines whether these relevance shifts are linked to hallucination behavior. It first compares relevance changes across samples with different prediction outcomes, and then tests how behavior changes when relevance is increased or decreased for different token groups.

In the vision setting, we follow the POPE benchmark protocol and randomly sample 100 examples with a positive label, using the bounding box of the queried object as the annotated relevant region. For audio, we analogously sample 100 positive examples from DCASE 2019 Task 4~\citep{turpault2019sound}, generate queries of the form ``Is [sound event] present in the audio?'', and use the annotated temporal segment of the queried event as the relevant region. For each sample, we generate responses using both the baseline and our method and compute token-level relevance scores with respect to the predicted answer, normalized to $[0,1]$ across samples.

Following the notation in Eq.~\eqref{eq:totalRelevance}, let $\Phi_{\mathcal{P}}$ and $\Phi_{\mathcal{T}}$ denote the total relevance assigned to perceptual and textual tokens, respectively. Using the query-relevant perceptual region $\mathcal{G}$ defined in Section~\ref{sec:inferenceTimeOptimization}, we define \emph{relevance localization} as the fraction of perceptual relevance localized within the ground-truth region, given by $\sum_{i\in\mathcal{G}}\Phi_i/\Phi_{\mathcal{P}}$. Since this quantity is normalized by $\Phi_{\mathcal{P}}$, it is invariant to a uniform rescaling of relevance across perceptual tokens. In addition, we define \emph{perceptual utilization} as $\Phi_{\mathcal{P}}/(\Phi_{\mathcal{P}}+\Phi_{\mathcal{T}})$, measuring the relative contribution of perceptual evidence to the prediction. For relevance localization, we report two configurations of the full inference-time objective in Eq.~\eqref{eq:objective}: one using our annotation-free relevance loss in Eq.~\eqref{eq:relevanceLoss}, and one using the supervised relevance loss in Eq.~\eqref{eq:supervisedrelevanceLoss}, which we denote \emph{LIME-Sup}. In both cases, the same KL regularization is applied. We evaluate each at three, five, and seven optimization steps to observe how the redistribution evolves. As stated in Section~\ref{sec:inferenceTimeOptimization}, LIME does not assume knowledge of $\mathcal{G}$, which is used here for analysis only, and since LIME-Sup targets the same region over which relevance localization is measured, it serves as an upper bound rather than a deployable configuration.

The results are presented in Table~\ref{tab:modalityUtilization}. We observe that our approach consistently improves both relevance localization and perceptual utilization across vision and audio models. In particular, the increase in perceptual utilization indicates that the model assigns greater explanatory responsibility to perceptual tokens, while the improvement in relevance localization suggests that this relevance becomes more concentrated on semantically relevant regions. The redistribution within $\mathcal{P}$ is therefore selective, favoring the query-relevant region even though the objective does not identify it. Comparing the two configurations quantifies how selective it is, LIME-Sup is stronger at every budget, but the gap narrows as optimization proceeds: by seven steps our objective attains $81\%$ and $95\%$ of its improvement over vanilla decoding on the two models.  The annotation-free objective therefore converges toward the region-supervised one rather than plateauing at a coarser solution, without access to $\mathcal{G}$. Together, these results show that LIME not only shifts relevance from textual to perceptual tokens, but also selectively concentrates perceptual relevance within query-relevant regions, despite having no access to $\mathcal{G}$. As noted earlier, Appendix~\ref{app:mechanism} provides a local analysis of why this selective localization can emerge despite the objective having no access to the query-relevant regions.

\begin{figure}[t]
  \centering
  \includegraphics[width=\linewidth]{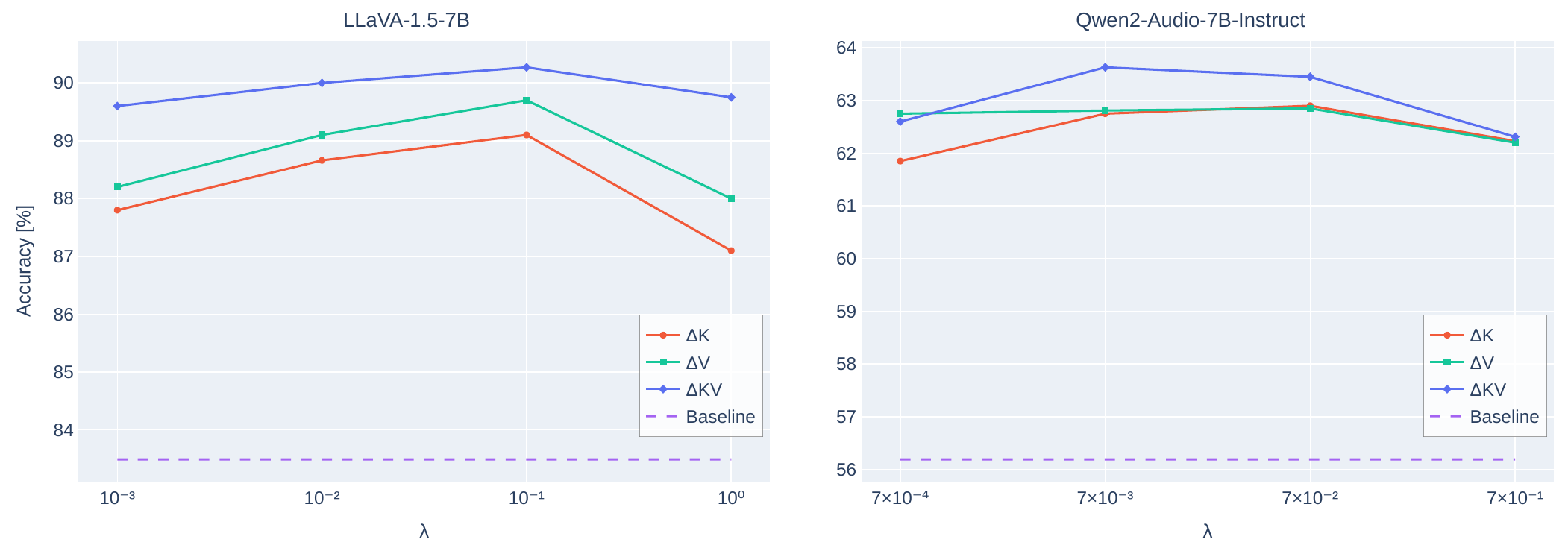}
  \caption{Effect of KL regularization $\lambda$ under different editing strategies. Here, $\Delta \vect{K}$ and $\Delta \vect{V}$ denote optimizing only the key or value perturbations, respectively, while $\Delta \vect{KV}$ denotes their joint optimization and corresponds to $\Delta$ in the main text. Results are shown for LLaVA-1.5-7B (left) and Qwen2-Audio-7B-Instruct (right).}
  \label{fig:klEffect}
\end{figure}

\subsection{Ablation Studies}
\label{sec:ablationStudies}

We next examine the main design choices of LIME by studying the roles of $\Delta \mathbf{K}$ and $\Delta \mathbf{V}$, and the sensitivity to the KL-regularization weight $\lambda$. Efficiency ablations varying the number of optimization steps and the layers to which $\Delta$ is applied, together with latency and memory measurements, are reported in Appendix~\ref{app:efficiency}. Unless otherwise stated, the vision experiments use LLaVA-1.5-7B on the random split of POPE-MSCOCO, while the audio experiments use Qwen2-Audio-7B-Instruct on the random split of AHQA.

\paragraph{Key--value updates and KL regularization.}
We first evaluate three variants of LIME: optimizing only the keys $\Delta \mathbf{K}$, only the values $\Delta \mathbf{V}$, or both jointly $\Delta $. For each variant, we sweep the KL-regularization weight $\lambda$, while fixing the number of optimization steps to seven. The results are presented in Figure~\ref{fig:klEffect}. All variants outperform the baseline, indicating that key and value updates independently contribute to improving multimodal grounding. Among them, jointly modifying both consistently achieves the best performance, indicating that these components play complementary roles in controlling attention
and information aggregation. Smaller values of $\lambda$ enable stronger updates to the model's internal representations, while larger values constrain the optimization and reduce its effect. Across both domains, intermediate values of $\lambda$ provide the best trade-off between enhancing modality utilization and preserving the model's original behavior.

\section{Conclusion}
\label{sec:conclusion}

We studied multimodal hallucinations through the lens of modality utilization at inference time. Using LRP, we directly characterize the contribution of perceptual and textual tokens, showing that hallucinations are associated with reduced perceptual relevance and that intervening on this relevance systematically changes prediction behavior. Building on this analysis, we introduced LIME, a training-free framework that uses the model's own relevance signal to optimize key-value representations during decoding without modifying model parameters or requiring training data. Across vision and audio benchmarks, LIME consistently reduces hallucinations while preserving generation quality. Moreover, despite receiving no spatial or temporal supervision, LIME increases perceptual utilization and concentrates relevance on query-relevant regions, demonstrating that relevance provides an effective inference-time signal for controlling multimodal generation.

LIME incurs additional computational overhead due to per-step optimization during decoding, as discussed in detail in Appendix~\ref{app:limitations}. Nevertheless, the optimization budget and intervention scope can be adjusted to obtain lower-cost variants while retaining meaningful gains. The method also requires hyperparameter selection, which may limit its applicability in latency-sensitive or resource-constrained settings.

\section*{AI use statement}
In this work, we used generative AI tools solely to improve the presentation of the manuscript, including grammar, readability, and clarity. All research ideas, methodological developments, experiments, and analyses were conceived and conducted by the authors. We take responsibility for the final content of this work,
including text, claims or artifacts produced with the aid of generative AI.

\bibliography{ref}

@inproceedings{liu2024paying,
  title={Paying more attention to image: A training-free method for alleviating hallucination in lvlms},
  author={Liu, Shi and Zheng, Kecheng and Chen, Wei},
  booktitle={European Conference on Computer Vision},
  pages={125--140},
  year={2024},
  organization={Springer}
}

@article{zoulook,
  title={Look Twice Before You Answer: Memory-Space Visual Retracing for Hallucination Mitigation in Multimodal Large Language Models}, 
  author={Zou, Xin and Wang, Yizhou and Yan, Yibo and Lyu, Yuanhuiyi and Zheng, Kening and Huang, Sirui and Chen, Junkai and Jiang, Peijie and Liu, Jia and Tang, Chang and Hu, Xuming},
  journal={Forty-second International Conference on Machine Learning (ICML)},
  year={2025}
}

@article{sun2025v,
  title={V-ITI: Mitigating Hallucinations in Multimodal Large Language Models via Visual Inference-Time Intervention},
  author={Sun, Nan and Zhang, Zhenyu and Lin, Xixun and Wang, Kun and Shang, Yanmin and Gu, Naibin and Wang, Shuohuan and Sun, Yu and Wu, Hua and Wang, Haifeng and others},
  journal={arXiv preprint arXiv:2512.03542},
  year={2025}
}

@article{hsu2025reducing,
  title={Reducing object hallucination in large audio-language models via audio-aware decoding},
  author={Hsu, Tzu-wen and Lu, Ke-Han and Chiang, Cheng-Han and Lee, Hung-yi},
  journal={arXiv preprint arXiv:2506.07233},
  year={2025}
}

@article{chu2024qwen2,
  title={Qwen2-audio technical report},
  author={Chu, Yunfei and Xu, Jin and Yang, Qian and Wei, Haojie and Wei, Xipin and Guo, Zhifang and Leng, Yichong and Lv, Yuanjun and He, Jinzheng and Lin, Junyang and others},
  journal={arXiv preprint arXiv:2407.10759},
  year={2024}
}

@article{Qwen-VL,
  title={Qwen-VL: A Versatile Vision-Language Model for Understanding, Localization, Text Reading, and Beyond},
  author={Bai, Jinze and Bai, Shuai and Yang, Shusheng and Wang, Shijie and Tan, Sinan and Wang, Peng and Lin, Junyang and Zhou, Chang and Zhou, Jingren},
  journal={arXiv preprint arXiv:2308.12966},
  year={2023}
}

@article{Qwen2.5-VL,
  title={Qwen2.5-VL Technical Report},
  author={Bai, Shuai and Chen, Keqin and Liu, Xuejing and Wang, Jialin and Ge, Wenbin and Song, Sibo and Dang, Kai and Wang, Peng and Wang, Shijie and Tang, Jun and Zhong, Humen and Zhu, Yuanzhi and Yang, Mingkun and Li, Zhaohai and Wan, Jianqiang and Wang, Pengfei and Ding, Wei and Fu, Zheren and Xu, Yiheng and Ye, Jiabo and Zhang, Xi and Xie, Tianbao and Cheng, Zesen and Zhang, Hang and Yang, Zhibo and Xu, Haiyang and Lin, Junyang},
  journal={arXiv preprint arXiv:2502.13923},
  year={2025}
}

@article{liu2023visual,
  title={Visual instruction tuning},
  author={Liu, Haotian and Li, Chunyuan and Wu, Qingyang and Lee, Yong Jae},
  journal={Advances in neural information processing systems},
  volume={36},
  pages={34892--34916},
  year={2023}
}

@inproceedings{
  tang2024salmonn,
  title={SALMONN: Towards Generic Hearing Abilities for Large Language Models},
  author={Changli Tang and Wenyi Yu and Guangzhi Sun and Xianzhao Chen and Tian Tan and Wei Li and Lu Lu and Zejun MA and Chao Zhang},
  booktitle={The Twelfth International Conference on Learning Representations},
  year={2024},
  url={https://openreview.net/forum?id=14rn7HpKVk}
}

@article{liu2024survey,
  title={A survey on hallucination in large vision-language models},
  author={Liu, Hanchao and Xue, Wenyuan and Chen, Yifei and Chen, Dapeng and Zhao, Xiutian and Wang, Ke and Hou, Liping and Li, Rongjun and Peng, Wei},
  journal={arXiv preprint arXiv:2402.00253},
  year={2024}
}

@inproceedings{huang2024visual,
  title={Visual hallucinations of multi-modal large language models},
  author={Huang, Wen and Liu, Hongbin and Guo, Minxin and Gong, Neil},
  booktitle={Findings of the Association for Computational Linguistics: ACL 2024},
  pages={9614--9631},
  year={2024}
}

@inproceedings{zheng2025reefknot,
  title={Reefknot: A comprehensive benchmark for relation hallucination evaluation, analysis and mitigation in multimodal large language models},
  author={Zheng, Kening and Chen, Junkai and Yan, Yibo and Zou, Xin and Zhou, Huiyu and Hu, Xuming},
  booktitle={Findings of the Association for Computational Linguistics: ACL 2025},
  pages={6193--6212},
  year={2025}
}

@inproceedings{kuan2025can,
  title={Can large audio-language models truly hear? tackling hallucinations with multi-task assessment and stepwise audio reasoning},
  author={Kuan, Chun-Yi and Lee, Hung-yi},
  booktitle={ICASSP 2025-2025 IEEE International Conference on Acoustics, Speech and Signal Processing (ICASSP)},
  pages={1--5},
  year={2025},
  organization={IEEE}
}

@article{bach2015pixel,
  title={On pixel-wise explanations for non-linear classifier decisions by layer-wise relevance propagation},
  author={Bach, Sebastian and Binder, Alexander and Montavon, Gr{\'e}goire and Klauschen, Frederick and M{\"u}ller, Klaus-Robert and Samek, Wojciech},
  journal={PloS one},
  volume={10},
  number={7},
  pages={e0130140},
  year={2015},
  publisher={Public Library of Science San Francisco, CA USA}
}

@article{montavon2017explaining,
  title={Explaining nonlinear classification decisions with deep taylor decomposition},
  author={Montavon, Gr{\'e}goire and Lapuschkin, Sebastian and Binder, Alexander and Samek, Wojciech and M{\"u}ller, Klaus-Robert},
  journal={Pattern recognition},
  volume={65},
  pages={211--222},
  year={2017},
  publisher={Elsevier}
}

@InProceedings{pmlr-v235-achtibat24a,
  title = {{A}ttn{LRP}: Attention-Aware Layer-Wise Relevance Propagation for Transformers},
  author = {Achtibat, Reduan and Hatefi, Sayed Mohammad Vakilzadeh and Dreyer, Maximilian and Jain, Aakriti and Wiegand, Thomas and Lapuschkin, Sebastian and Samek, Wojciech},
  booktitle = {Proceedings of the 41st International Conference on Machine Learning},
  pages = {135--168},
  year = {2024},
  editor = {Salakhutdinov, Ruslan and Kolter, Zico and Heller, Katherine and Weller, Adrian and Oliver, Nuria and Scarlett, Jonathan and Berkenkamp, Felix},
  volume = {235},
  series = {Proceedings of Machine Learning Research},
  month = {21--27 Jul},
  publisher = {PMLR}
}

@article{touvron2023llama,
  title={Llama 2: Open foundation and fine-tuned chat models},
  author={Touvron, Hugo and Martin, Louis and Stone, Kevin and Albert, Peter and Almahairi, Amjad and Babaei, Yasmine and Bashlykov, Nikolay and Batra, Soumya and Bhargava, Prajjwal and Bhosale, Shruti and others},
  journal={arXiv preprint arXiv:2307.09288},
  year={2023}
}

@article{bai2023qwen,
  title={Qwen technical report},
  author={Bai, Jinze and Bai, Shuai and Chu, Yunfei and Cui, Zeyu and Dang, Kai and Deng, Xiaodong and Fan, Yang and Ge, Wenbin and Han, Yu and Huang, Fei and others},
  journal={arXiv preprint arXiv:2309.16609},
  year={2023}
}

@inproceedings{gunjal2024detecting,
  title={Detecting and preventing hallucinations in large vision language models},
  author={Gunjal, Anisha and Yin, Jihan and Bas, Erhan},
  booktitle={Proceedings of the AAAI Conference on Artificial Intelligence},
  volume={38 (16)},
  pages={18135--18143},
  year={2024}
}

@inproceedings{yu2024rlhf,
  title={Rlhf-v: Towards trustworthy mllms via behavior alignment from fine-grained correctional human feedback},
  author={Yu, Tianyu and Yao, Yuan and Zhang, Haoye and He, Taiwen and Han, Yifeng and Cui, Ganqu and Hu, Jinyi and Liu, Zhiyuan and Zheng, Hai-Tao and Sun, Maosong and others},
  booktitle={Proceedings of the IEEE/CVF Conference on Computer Vision and Pattern Recognition},
  pages={13807--13816},
  year={2024}
}

@inproceedings{zhou2024analyzing,
  title={Analyzing and mitigating object hallucination in large vision-language models},
  author={Zhou, Yiyang and Cui, Chenhang and Yoon, Jaehong and Zhang, Linjun and Deng, Zhun and Finn, Chelsea and Bansal, Mohit and Yao, Huaxiu},
  booktitle={International Conference on Learning Representations},
  volume={2024},
  pages={56969--56998},
  year={2024}
}

@inproceedings{huang2024opera,
  title={Opera: Alleviating hallucination in multi-modal large language models via over-trust penalty and retrospection-allocation},
  author={Huang, Qidong and Dong, Xiaoyi and Zhang, Pan and Wang, Bin and He, Conghui and Wang, Jiaqi and Lin, Dahua and Zhang, Weiming and Yu, Nenghai},
  booktitle={Proceedings of the IEEE/CVF Conference on Computer Vision and Pattern Recognition},
  pages={13418--13427},
  year={2024}
}

@inproceedings{li2023contrastive,
  title={Contrastive decoding: Open-ended text generation as optimization},
  author={Li, Xiang Lisa and Holtzman, Ari and Fried, Daniel and Liang, Percy and Eisner, Jason and Hashimoto, Tatsunori B and Zettlemoyer, Luke and Lewis, Mike},
  booktitle={Proceedings of the 61st annual meeting of the association for computational linguistics (volume 1: Long papers)},
  pages={12286--12312},
  year={2023}
}

@inproceedings{leng2024mitigating,
  title={Mitigating object hallucinations in large vision-language models through visual contrastive decoding},
  author={Leng, Sicong and Zhang, Hang and Chen, Guanzheng and Li, Xin and Lu, Shijian and Miao, Chunyan and Bing, Lidong},
  booktitle={Proceedings of the IEEE/CVF Conference on Computer Vision and Pattern Recognition},
  pages={13872--13882},
  year={2024}
}

@inproceedings{wang2024mitigating,
  title={Mitigating hallucinations in large vision-language models with instruction contrastive decoding},
  author={Wang, Xintong and Pan, Jingheng and Ding, Liang and Biemann, Chris},
  booktitle={Findings of the Association for Computational Linguistics: ACL 2024},
  pages={15840--15853},
  year={2024}
}

@inproceedings{kuan2024understanding,
  title={Understanding Sounds, Missing the Questions: The Challenge of Object Hallucination in Large Audio-Language Models},
  author={Kuan, Chun-Yi and Huang, Wei-Ping and Lee, Hung-yi},
  booktitle={Proc of International Speech Communication Association (INTERSPEECH)},
  year={2024},
}

@inproceedings{yang2024air,
  title={Air-bench: Benchmarking large audio-language models via generative comprehension},
  author={Yang, Qian and Xu, Jin and Liu, Wenrui and Chu, Yunfei and Jiang, Ziyue and Zhou, Xiaohuan and Leng, Yichong and Lv, Yuanjun and Zhao, Zhou and Zhou, Chang and others},
  booktitle={Proceedings of the 62nd Annual Meeting of the Association for Computational Linguistics (Volume 1: Long Papers)},
  pages={1979--1998},
  year={2024}
}

@inproceedings{rohrbach-etal-2018-object,
    title = "Object Hallucination in Image Captioning",
    author = "Rohrbach, Anna  and
      Hendricks, Lisa Anne  and
      Burns, Kaylee  and
      Darrell, Trevor  and
      Saenko, Kate",
    editor = "Riloff, Ellen  and
      Chiang, David  and
      Hockenmaier, Julia  and
      Tsujii, Jun{'}ichi",
    booktitle = "Proceedings of the 2018 Conference on Empirical Methods in Natural Language Processing",
    month = oct # "-" # nov,
    year = "2018",
    address = "Brussels, Belgium",
    publisher = "Association for Computational Linguistics",
    url = "https://aclanthology.org/D18-1437/",
    doi = "10.18653/v1/D18-1437",
    pages = "4035--4045",
}

@inproceedings{lin2014microsoft,
  title={Microsoft coco: Common objects in context},
  author={Lin, Tsung-Yi and Maire, Michael and Belongie, Serge and Hays, James and Perona, Pietro and Ramanan, Deva and Doll{\'a}r, Piotr and Zitnick, C Lawrence},
  booktitle={European conference on computer vision},
  pages={740--755},
  year={2014},
  organization={Springer}
}

@inproceedings{schwenk2022okvqa,
  title={A-okvqa: A benchmark for visual question answering using world knowledge},
  author={Schwenk, Dustin and Khandelwal, Apoorv and Clark, Christopher and Marino, Kenneth and Mottaghi, Roozbeh},
  booktitle={European conference on computer vision},
  pages={146--162},
  year={2022},
  organization={Springer}
}

@inproceedings{li2023evaluating,
  title={Evaluating object hallucination in large vision-language models},
  author={Li, Yifan and Du, Yifan and Zhou, Kun and Wang, Jinpeng and Zhao, Wayne Xin and Wen, Ji-Rong},
  booktitle={Proceedings of the 2023 conference on empirical methods in natural language processing},
  pages={292--305},
  year={2023}
}

@article{asadi2026mirage,
  title={Mirage The Illusion of Visual Understanding},
  author={Asadi, Mohammad and O'Sullivan, Jack W and Cao, Fang and Nedaee, Tahoura and Fardi, Kamyar and Li, Fei-Fei and Adeli, Ehsan and Ashley, Euan},
  journal={arXiv preprint arXiv:2603.21687},
  year={2026}
}

@inproceedings{turpault2019sound,
  title={Sound event detection in domestic environments with weakly labeled data and soundscape synthesis},
  author={Turpault, Nicolas and Serizel, Romain and Shah, Ankit Parag and Salamon, Justin},
  booktitle={Workshop on Detection and Classification of Acoustic Scenes and Events},
  year={2019}
}

@inproceedings{radford2021learning,
  title={Learning transferable visual models from natural language supervision},
  author={Radford, Alec and Kim, Jong Wook and Hallacy, Chris and Ramesh, Aditya and Goh, Gabriel and Agarwal, Sandhini and Sastry, Girish and Askell, Amanda and Mishkin, Pamela and Clark, Jack and others},
  booktitle={International conference on machine learning},
  pages={8748--8763},
  year={2021},
  organization={PmLR}
}

@inproceedings{radford2023robust,
  title={Robust speech recognition via large-scale weak supervision},
  author={Radford, Alec and Kim, Jong Wook and Xu, Tao and Brockman, Greg and McLeavey, Christine and Sutskever, Ilya},
  booktitle={International conference on machine learning},
  pages={28492--28518},
  year={2023},
  organization={PMLR}
}

@misc{vicuna2023,
    title = {Vicuna: An Open-Source Chatbot Impressing GPT-4 with 90\%* ChatGPT Quality},
    url = {https://lmsys.org/blog/2023-03-30-vicuna/},
    author = {Chiang, Wei-Lin and Li, Zhuohan and Lin, Zi and Sheng, Ying and Wu, Zhanghao and Zhang, Hao and Zheng, Lianmin and Zhuang, Siyuan and Zhuang, Yonghao and Gonzalez, Joseph E. and Stoica, Ion and Xing, Eric P.},
    month = {March},
    year = {2023}
}

@InProceedings{Elisha_2026_CVPR,
    author    = {Elisha, Yehonatan and Barkan, Oren and Koenigstein, Noam},
    title     = {Concept-Guided Fine-Tuning: Steering ViTs away from Spurious Correlations to Improve Robustness},
    booktitle = {Proceedings of the IEEE/CVF Conference on Computer Vision and Pattern Recognition (CVPR)},
    month     = {June},
    year      = {2026},
    pages     = {17035-17045}
}

@article{kingma2014adam,
  title={Adam: A method for stochastic optimization},
  author={Kingma, Diederik P and Ba, Jimmy},
  journal={arXiv preprint arXiv:1412.6980},
  year={2014}
}

@article{Qwen2.5-Omni,
  title={Qwen2.5-Omni Technical Report},
  author={Xu, Jin and Guo, Zhifang and He, Jinzheng and Hu, Hangrui and He, Ting and Bai, Shuai and Chen, Keqin and Wang, Jialin and Fan, Yang and Dang, Kai and Zhang, Bin and Wang, Xiong and Chu, Yunfei and Lin, Junyang},
  journal={arXiv preprint arXiv:2503.20215},
  year={2025}
}

@inproceedings{guan2024hallusionbench,
  title={Hallusionbench: an advanced diagnostic suite for entangled language hallucination and visual illusion in large vision-language models},
  author={Guan, Tianrui and Liu, Fuxiao and Wu, Xiyang and Xian, Ruiqi and Li, Zongxia and Liu, Xiaoyu and Wang, Xijun and Chen, Lichang and Huang, Furong and Yacoob, Yaser and others},
  booktitle={Proceedings of the IEEE/CVF conference on computer vision and pattern recognition},
  pages={14375--14385},
  year={2024}
}

@article{cui2026minicpm,
  title={Minicpm-o 4.5: Towards real-time full-duplex omni-modal interaction},
  author={Cui, Junbo and Xu, Bokai and Wang, Chongyi and Yu, Tianyu and Sun, Weiyue and Xu, Yingjing and Wang, Tianran and He, Zhihui and Ma, Wenshuo and Cai, Tianchi and others},
  journal={arXiv preprint arXiv:2604.27393},
  year={2026}
}

@inproceedings{guo2024llava,
  title={Llava-uhd: an lmm perceiving any aspect ratio and high-resolution images},
  author={Guo, Zonghao and Xu, Ruyi and Yao, Yuan and Cui, Junbo and Ni, Zanlin and Ge, Chunjiang and Chua, Tat-Seng and Liu, Zhiyuan and Huang, Gao},
  booktitle={European Conference on Computer Vision},
  pages={390--406},
  year={2024},
  organization={Springer}
}

@inproceedings{saunshi2022understanding,
  title={Understanding contrastive learning requires incorporating inductive biases},
  author={Saunshi, Nikunj and Ash, Jordan and Goel, Surbhi and Misra, Dipendra and Zhang, Cyril and Arora, Sanjeev and Kakade, Sham and Krishnamurthy, Akshay},
  booktitle={International Conference on Machine Learning},
  pages={19250--19286},
  year={2022},
  organization={PMLR}
}
\bibliographystyle{iclr2027_conference}

\appendix
\section*{Appendix}
\setcounter{table}{0}
\renewcommand{\thetable}{A.\arabic{table}}

\section{Layer-wise Relevance Propagation for Transformer Models}
\label{app:layerWiseRelevancePropogation}
This appendix provides additional details on the Layer-wise Relevance Propagation (LRP) framework used to compute token-level relevance scores in our method. We briefly review the relevance decomposition principle, the conservation property, and the transformer-specific propagation rules used throughout this work. Our formulation follows the Attention-Aware Layer-wise Relevance Propagation~\citep{pmlr-v235-achtibat24a} (AttnLRP) framework, from which the transformer-specific propagation rules presented below are adopted.

LRP is a class of attribution methods that explains a scalar model output by decomposing it into additive contributions of intermediate representations and input features. Consider scalar function $f_j$ with input vector $\vect{x}$ of size $N$. LRP assigns relevance values $\{\Phi_{i\leftarrow j}\}_{i=0}^{N-1}$, where $\Phi_{i\leftarrow j}$ reflects the amount of the output $f_j$ that is attributable to input $i$. These relevance values are defined such that their sum is proportional to the function output:
\begin{equation}
    \label{eq:4}
    f_j(\vect{x})\propto
    \Phi_{j}=
    \sum_{i}\Phi_{i\leftarrow j}.
\end{equation}    
When an input $i$ contributes to multiple neurons $j$, its total relevance, $\Phi_{i}$, is obtained by aggregating its contributions across all connected outputs:
\begin{equation}
    \label{eq:5}
    \Phi_{i}=\sum_j\Phi_{i\leftarrow j}.
\end{equation}
A central property of LRP is the \emph{conservation law}. When relevance is propagated backward through a layered network, the total relevance is preserved across layers. Denoting by $\Phi_{j}^{\ell}$ the relevance assigned to neuron $j$ at layer $\ell$, relevance propagation satisfies the following \emph{conservation law}:
\begin{equation}
    \label{eq:6}
    \Phi^{\ell-1}=
    \sum_i\Phi_{i}^{\ell - 1}=
    \sum_j\Phi_{j}^{\ell}=
    \Phi^{\ell}.
\end{equation}
This property ensures that relevance is neither created nor destroyed as it flows from the output layer toward the input. As a result, relevance values remain comparable across layers and retain a direct relationship to the original scalar output being explained.

Unlike gradient-based attribution methods, whose magnitudes may shrink or explode through depth, LRP maintains a relevance-preserving decomposition. To derive propagation rules, LRP is commonly formulated through Deep Taylor Decomposition (DTD)~\citep{montavon2017explaining}. The key idea is to locally approximate each neuron's computation by a first-order Taylor expansion around a reference point, and to interpret the resulting decomposition as additive relevance contributions. Consider a neuron $j$ represented as a scalar function $f_j(\vect{x})$ where $\vect{x}=(x_0,...,x_{N-1})$ denotes its input activations. DTD approximates this function locally by a first-order Taylor expansion around a reference point $\vect{x}^0=(x^0_0,...,x^0_{N-1})$
\[
    f_j(\vect{x})\approx
    f_j(\vect{x}^0)+\sum_i\frac{\partial f_j}{\partial x_i}(\vect{x}^0)(x_i-x^0_i).
\]
Rearranging terms yields an affine approximation of the form
\[
    f_j(\vect{x})\approx
    \sum_i \mathbf{J}_{ji}x_i+b^0_j,
\]

where $\mathbf{J}$ denotes the Jacobian evaluated at the reference point, and $b^0_j$ collects the bias and higher-order approximation error.

Under the additive relevance assumption~\citep{pmlr-v235-achtibat24a}, relevance is taken to be proportional to the neuron's output. Scaling the affine approximation by a constant factor $c\in \mathbb{R}$ with $f_j(\vect{x}) \neq 0$. Following Eq~\eqref{eq:4} yields
\begin{equation}
    \label{eq:7}
    \Phi_{j}=
    c f_j(\vect{x})=
    \sum_i \underbrace{\mathbf{J}_{ji}x_i \frac{\Phi_{j}}{f_j(\vect{x})}}_{\Phi_{i \leftarrow j}}
    + \underbrace{ b_j^0 \frac{\Phi_{j}}{f_j(\vect{x})}}_{\Phi_{b \leftarrow j}}.
\end{equation}
Comparing with Eq~\eqref{eq:4}, the first term in the decomposition corresponds to the relevance assigned to the input variables, while the second term captures the contribution of bias and higher-order approximation error. Following the AttnLRP formulation, we focus on propagating input-dependent relevance and omit the bias-related term, since it does not encode input-specific information and can be modeled as a constant contribution. This simplification preserves the conservation property~\eqref{eq:6} and does not affect the interpretation of token-level relevance. Applying Eq~\eqref{eq:5} and Eq~\eqref{eq:7} yields
\[
    \Phi_{i}=
    \sum_j \Phi_{i \leftarrow j}=
    \sum_j \mathbf{J}_{ji}x_i\frac{\Phi_{j}}{f_j(\vect{x})}.
\]
The expression above provides a general relevance decomposition derived from local linearization. For neural network layers, it can be written in a more practical form using activations and layer weights, resulting in the standard LRP-z rule:
\[
    \Phi_{j}^{\ell}=
    \sum_{i}\frac{x_{j}^{\ell} W_{ji}^{\ell}}{\sum_{k}x_{k}^{\ell} W_{ki}^{\ell}} \Phi_{i}^{\ell +1},
\]
where $\vect{x}^{\ell}$ denotes the activations at layer $\ell$, and $\mathbf{W}^{\ell}$ denotes the associated weights. The denominator normalizes the redistribution such that relevance conservation is preserved across layers. This rule forms the basis for propagating relevance through linear transformations.

While the LRP-z rule is suitable for many linear operations, transformer architectures contain additional components whose computations do not naturally follow simple linear redistribution. Consequently, specialized propagation rules are required to maintain stable and faithful attribution.

\paragraph{LRP-$\epsilon$ rule.}
The first extension is the LRP-$\epsilon$ rule, which stabilizes propagation when denominators become numerically small. In deep networks, the normalization term in the z-rule may approach zero, producing unstable relevance assignments. To address this, a small stabilizing term is introduced:
\[
    \Phi^{\ell}_j = 
    \sum_i \frac{x^{\ell}_j  W^{\ell}_{ji}}{\sum_k x^{\ell}_k  W^{\ell}_{ki} + \epsilon \,\mathrm{sign}\!\left(\sum_m x^{\ell}_m  W^{\ell}_{mi}\right)} \, \Phi^{\ell + 1}_i ,
\]
where $\epsilon > 0$ is a small constant. This stabilization prevents division by values close to zero while preserving the overall relevance decomposition. As a result, relevance propagation becomes more robust in deep transformer architectures where activations may vary substantially across layers.

\paragraph{LRP for softmax normalization.}
Beyond linear mappings, transformer models also include normalization operations that redistribute activations across dimensions. In particular, the softmax function introduces competition between neurons through normalization. Let
\[
    a_{j}^{\ell} = \mathrm{softmax}(\vect{x}^{\ell})_{j}=
    \frac{\exp(x_{j}^{\ell})}{\sum_k \exp(x_k^{\ell})}.
\]
Relevance propagation through softmax cannot be treated as an identity operation because each output depends on all input dimensions. Instead, relevance is redistributed according to
\[
    \Phi_{j}^{\ell - 1} =
    x_{j}^{\ell} \left( \Phi_{j}^{\ell} - a_{j}^{\ell} \sum_i \Phi_{i}^{\ell} \right).
\]
This formulation captures the competitive normalization behavior of softmax while enabling stable relevance propagation across dimensions.

\paragraph{LRP for matrix decomposition in attention.}
For the attention-value interaction, relevance must be propagated through the matrix product between attention weights and value representations. Let
\[
    \mathbf{O}^{\ell} = \mathbf{A}^{\ell} \mathbf{V}^{\ell},
\]
where $\mathbf{A}^{\ell}$ denotes the attention matrix and $\mathbf{V}^{\ell}$ the value tensor of layer $\ell$. Relevance propagation through this bilinear interaction is decomposed across the matrix product. The relevance assigned to attention entry $A_{ji}^{\ell}$ is given by
\[
    \Phi_{ji}^{\ell-1}
    =
    \sum_p
    \frac{
    A^{\ell}_{ji} V^{\ell}_{ip}
    }{
    2 O^{\ell}_{jp} + \epsilon
    }
    \Phi_{jp}^{\ell}.
\]
An analogous rule is applied to the value branch by summing over the complementary dimension. This decomposition preserves conservation while enabling attribution through attention-value interactions.

\paragraph{LRP for LayerNorm and RMSNorm.}
Finally, transformer architectures contain normalization layers such as LayerNorm and RMSNorm. These operations primarily rescale activations rather than redistribute information across features. Consequently, relevance propagation through normalization layers is approximated using an identity mapping
\[
    \Phi_{i}^{\ell-1} = \Phi_{i}^{\ell}.
\]
This approximation assumes that normalization preserves feature attribution while only modifying activation scale, allowing relevance to pass unchanged through these layers.

Together, these propagation rules provide a complete relevance decomposition framework for transformer architectures. By combining stabilized linear propagation, normalization-aware redistribution, and attention-specific matrix decomposition, relevance can be propagated consistently from output predictions back to multimodal input tokens.

\section{Experimental Setup}
\label{app:experimentalSetup}

\subsection{Evaluation Benchmarks}
\label{app:evaluationBenchmarks}

We provide additional details on the datasets and evaluation procedures used in our experiments. For each benchmark, we specify the prompting strategy and decoding parameters, including the maximum generation length. Unless stated otherwise, all methods are evaluated under identical prompting and decoding settings.

\paragraph{POPE~\citep{li2023evaluating}} Polling-based Object Probing Evaluation (POPE) is a benchmark designed to measure object hallucination in large vision–language models through a binary visual question answering task. Given an image, models are asked yes/no questions of the form "Is there a [object] in the image?". Positive questions correspond to objects that are present in the scene according to ground-truth annotations, while negative questions query objects that do not appear in the image. Following prior works~\citep{liu2024paying, zoulook, sun2025v}, we conduct the evaluation on both the MSCOCO dataset~\citep{lin2014microsoft} and the A-OKVQA dataset~\citep{schwenk2022okvqa}, using  500 randomly sampled images for each dataset, with multiple object queries. Similar to the official POPE protocol, negative queries are constructed using three sampling strategies that vary in difficulty; random, which samples objects uniformly from the object vocabulary; popular, which samples from frequently occurring objects in the dataset; and adversarial, which selects objects that commonly co-occur with the ground-truth objects in similar scenes, making them more challenging negatives. For each query, the model predicts whether the object is present in the image, and performance is evaluated using accuracy and F1 score. We report results separately for each sampling strategy, as well as the average across the three settings. To reduce ambiguity in free-form generation, we append the instruction "Please answer yes or no." to each prompt. We use a maximum generation length of 20 tokens. Final predictions are obtained following the standard POPE evaluation protocol.

\paragraph{CHAIR~\citep{rohrbach-etal-2018-object}}
Caption Hallucination Assessment with Image Relevance (CHAIR), evaluates object hallucination in image captioning models by measuring whether generated captions mention objects that are not present in the image. Following prior work~\citep{liu2024paying, zoulook, sun2025v}, we conduct the evaluation on the MSCOCO~\citep{lin2014microsoft} 2014 validation split and randomly sample 500 images for evaluation. For each image, the model is prompted to generate a descriptive caption using a generic instruction of the form "Please describe this image in detail.", following prior work as well. The generated captions are then analyzed using the CHAIR evaluation procedure. Ground-truth objects are obtained from MSCOCO instance annotations and reference captions, allowing the evaluator to determine which objects are truly present in the scene. CHAIR reports per-instance evaluation CHAIR$_I$ and per-sentence evaluation CHAIR$_S$, defined as
follows:
\[
    \text{CHAIR}_{I}=
    \frac{|\{\text{hallucinated objects}\}|}{|\{\text{all objects mentioned}\}|}
\]
\[
    \text{CHAIR}_{S}=
    \frac{|\{\text{sentences with hallucinated object}\}|}{|\{\text{all sentences}\}|}
\]
We use a maximum generation length of 150 tokens.

\paragraph{HallusionBench~\citep{guan2024hallusionbench}}
HallusionBench evaluates hallucination in diverse image-context reasoning scenarios, including OCR, charts, tables, mathematical figures, maps, and visual illusions. Following the official evaluation protocol, we report figure-level accuracy (fACC), question-level accuracy (qACC), and accuracy on the easy (easyA) and hard (hardA) subsets.

\paragraph{AIR-Bench~\citep{yang2024air}}
The AIR-Bench Foundation benchmark consists of 19 single-task abilities\ spanning three audio domains: speech, natural sounds, and music, totaling over 19k single-choice questions. Although the questions are presented in a multiple-choice format, models are required to generate answers freely, making the evaluation fully generative. We follow the official AIR-Bench evaluation protocol and use the GPT-4-based evaluator to determine correctness. The evaluator compares the generated hypothesis with the reference answer given the question and associated meta-information, and outputs a binary judgment. We report accuracy, averaged per domain (Speech, Sound, Music). We use a maximum generation length of 50 tokens.

\paragraph{Audio Hallucination QA~\citep{kuan2024understanding}}
The Audio Hallucination QA benchmark evaluates object hallucination in large audio-language models by formulating the task as a binary discriminative question-answering problem. Given an audio clip, models are asked yes/no questions of the form "Is there a sound of [object] in the audio?". Positive questions are constructed from ground-truth sound labels, while negative questions are generated using three sampling strategies; random, popular, and adversarial, designed to probe the model's susceptibility to hallucinating non-existent objects. In our evaluation, we report accuracy and F1 score under each of the three sampling strategies. These metrics directly measure the model's tendency to over-predict object presence and provide a focused assessment of hallucination behavior under discriminative settings. To reduce ambiguity in generation, we append the instruction "Please answer yes or no." to each prompt and use a maximum generation length of 20 tokens.

\subsection{Baselines and Implementation Details}
\label{app:baselinesAndImplementationDetails}

We provide additional implementation details for the proposed method. Unless stated otherwise, all hyperparameters are reported in Table~\ref{tab:A:1}. At each decoding step, we optimize the key and value updates using a small number of gradient-based updates, following the formulation in Section~\ref{sec:inferenceTimeOptimization}. We use the Adam optimizer~\citep{kingma2014adam} for all inference-time optimization steps. To improve stability and reduce the number of optimization variables, the updates are shared across all attention heads within each layer. After each decoding step, the perturbations are reset and re-initialized for the next token.

\begin{table}[h]
    \caption{Hyperparameters of the proposed method for each evaluated model.}
    \label{tab:A:1}
    \centering
    \resizebox{\linewidth}{!}{
        \begin{tabular}{lcccc}
        \toprule
        \textbf{Model} & \textbf{Optimization Steps} & \textbf{Learning Rate} & \textbf{KL Weight ($\lambda$)} & \textbf{Temperature ($\tau$)}\\
        \midrule
        LLaVA-1.5-7B & 7 & $3 \times 10^{-4}$ & $0.1$ & $0.1$ \\
        Qwen-VL-Chat & 7 & $4 \times 10^{-4}$ & 0.1 & $0.1$ \\
        Qwen2.5-VL-7B-Instruct & 7 & $3 \times 10^{-4}$ & 0.1 & $0.1$ \\
        SALMONN-7B & 7 & $3 \times 10^{-4}$ & 0.1 & $0.1$ \\
        Qwen2-Audio-7B-Instruct & 7 & $5 \times 10^{-4}$ & $7 \times 10^{-3}$ & $0.1$ \\
        Qwen2.5-Omni-7B & 7 & $4 \times 10^{-4}$ & $0.1$ & $0.1$ \\
        MiniCPM-o 4.5 & 7 & $5 \times 10^{-4}$ & $0.1$ & $0.1$ \\            
        \bottomrule
        \end{tabular}
    }
\end{table}

\subsection{Additional Experiments and Results}
\label{app:additionalExperimentsAndResults}

We provide additional experimental results and qualitative examples that complement the main findings presented in Section~\ref{sec:experiments}. These include extended evaluations across models and benchmarks, as well as visualizations illustrating the effect of our method on multimodal grounding. Together, these results further support the effectiveness of the proposed approach.

\begin{table}[h]
    \caption{Evaluation results on the POPE benchmark with Qwen-VL-Chat and Qwen2.5-VL-7B Instruct.}
    \label{tab:A:visionPope}
    \centering
    \resizebox{\linewidth}{!}{
        \begin{tabular}{llcccccccc}
        \toprule
        \multirow{2}{*}{\textbf{Dataset}} & \multirow{2}{*}{\textbf{Methods}} 
        & \multicolumn{2}{c}{\textbf{Random (\%)} $\uparrow$} 
        & \multicolumn{2}{c}{\textbf{Popular (\%)} $\uparrow$} 
        & \multicolumn{2}{c}{\textbf{Adversarial (\%)} $\uparrow$}
        & \multicolumn{2}{c}{\textbf{Average (\%)} $\uparrow$} \\
        \cmidrule(lr){3-4}
        \cmidrule(lr){5-6}
        \cmidrule(lr){7-8}
        \cmidrule(lr){9-10}
         &  & Acc & F1 & Acc & F1 & Acc & F1 & Acc & F1 \\
        \midrule
        \multirow{5}{*}{MSCOCO}
        & Qwen-VL-Chat & 84.73 & 82.67 & 84.13 & 82.06 & 82.26 & 80.37 & 83.7 & 81.7 \\
        & + VCD & 88.63 & 87.81 & 87.12 & 86.4 & 84.26 & 83.9 & 86.67 & 86.03 \\
        & + LIME (ours) & \textbf{89.1} & \textbf{88.42} & \textbf{88.12} & \textbf{87.89} & \textbf{84.55} & \textbf{84.78} & \textbf{87.25} & \textbf{87.03} \\
        \cmidrule(lr){2-10}
        & Qwen2.5-VL-7B-Instruct & 86.6 & 84.48 & 85.2 & 83.12 & 84.23 & 82.16 & 85.34 & 83.25 \\
        & + LIME (ours) & \textbf{88.11} & \textbf{86.32} & \textbf{85.77} & \textbf{84.1} & \textbf{85.33} & \textbf{83.88} & \textbf{86.4} & \textbf{84.76} \\
        
        \midrule
        \multirow{5}{*}{A-OKVQA}
        & Qwen-VL-Chat & 86.67 & 85.59 & 85.56 & 84.63 & 79.57 & 79.5 & 83.93 & 83.24 \\
        & + VCD & \textbf{89.22} & \textbf{89.01} & 87.85 & 87.81 & 81.27 & 82.38 & 86.11 & \textbf{86.4} \\
        & + LIME (ours) & 87.8 & 87.21 & \textbf{87.95} & \textbf{87.95} & \textbf{83.21} & \textbf{83.36} & \textbf{86.32} & 86.17 \\        
        \cmidrule(lr){2-10}
        & Qwen2.5-VL-7B-Instruct & 88.53 & 87.23 & 86.12 & 85.07 & 85.05 & 84.14 & 86.56 & 85.48 \\
        & + LIME (ours) & \textbf{89.97} & \textbf{88.73} & \textbf{87.22} & \textbf{86.52} & \textbf{86.22} & \textbf{86} & \textbf{87.8} & \textbf{87.08} \\          
        \bottomrule    
        \end{tabular}
    }
\end{table}

\begin{table}[h]
    \caption{POPE benchmark results using LLaVA-1.5-7B on the A-OKVQA dataset. We report Accuracy and F1 for the Random, Popular, and Adversarial splits, as well as their average. Best and second-best results are highlighted in \textbf{bold} and \underline{underlined}, respectively.}
    \label{tab:A:3}
    \centering
    \resizebox{\linewidth}{!}{
        \begin{tabular}{llcccccccc}
        \toprule
        \multirow{2}{*}{\textbf{Dataset}} & \multirow{2}{*}{\textbf{Methods}}
        & \multicolumn{2}{c}{\textbf{Random (\%)} $\uparrow$} 
        & \multicolumn{2}{c}{\textbf{Popular (\%)} $\uparrow$} 
        & \multicolumn{2}{c}{\textbf{Adversarial (\%)} $\uparrow$}
        & \multicolumn{2}{c}{\textbf{Average (\%)} $\uparrow$} \\
        \cmidrule(lr){3-4}
        \cmidrule(lr){5-6}
        \cmidrule(lr){7-8}
        \cmidrule(lr){9-10}
        & & Acc & F1 & Acc & F1 & Acc & F1 & Acc & F1 \\
        \midrule
        \multirow{7}{*}{A-OKVQA}
        & LLaVA-1.5-7B & 83.45 & 82.56 & 79.9 & 79.59 & 74.04 & 75.15 & 79.13 & 79.1 \\
        & + OPERA & 88.27 & 87.54 & 85.17 & 84.74 & 79.37 & 79.97 & 84.27 & 84.08 \\
        & + VCD & 86.15 & 86.34 & 81.85 & 82.82 & 74.97 & 77.73 & 80.99 & 82.3 \\
        & + ICD & 85.57 & 85.06 & 81.93 & 81.95 & 77.43 & 78.99 & 81.64 & 82 \\
        & + MemVR & 91.1 & 90.83 & 87.33 & \underline{87.43} & 80.20 & \underline{81.66} & 86.21 & \underline{86.64} \\
        & + V-ITI & \underline{91.51} & \underline{91.04} & \underline{87.47} & 86.97 & \underline{80.34} & 81.12 & \underline{86.44} & 86.37 \\
        & + LIME (ours) & \textbf{91.71} & \textbf{91.12} & \textbf{88.75} & \textbf{87.93} & \textbf{81.55} & \textbf{82.4} & \textbf{87.33} & \textbf{87.15} \\
        \bottomrule    
        \end{tabular}
    }
\end{table}

\begin{table}[t]
    \centering
    \caption{Evaluation results on the random splits of POPE, using the MSCOCO and A-OKVQA datasets, and Audio Hallucination QA. Accuracy and F1 scores are reported for all settings.}
    \label{tab:A:omniPope}
    \setlength{\tabcolsep}{5pt}
    \begin{tabular}{lcccccc}
        \toprule
        \multirow{3}{*}{Methods}
        & \multicolumn{4}{c}{\textbf{POPE} \textbf{(\%) $\uparrow$}}
        & \multicolumn{2}{c}{
            \multirow{2}{*}{
                \shortstack{
                    \textbf{AHQA (\%) $\uparrow$}
                }
            }
        } \\
        \cmidrule(lr){2-5}
        & \multicolumn{2}{c}{MSCOCO}
        & \multicolumn{2}{c}{A-OKVQA}
        & \multicolumn{2}{c}{} \\
        \cmidrule(lr){2-3}
        \cmidrule(lr){4-5}
        \cmidrule(lr){6-7}
        & Acc & F1
        & Acc & F1
        & Acc & F1 \\
        \midrule
        Qwen2.5-Omni-7B
        & 87.7 & 86
        & 88.1 & 87.9
        & 89.25 & 89.34 \\
        + LIME (ours)
        & \textbf{89.5} & \textbf{88.12}
        & \textbf{89.5} & \textbf{89.3}
        & \textbf{90.5} & \textbf{90.33} \\
        \bottomrule
    \end{tabular}
\end{table}

\begin{table}[t]
    \centering
    \caption{Evaluation results on the random splits using the MSCOCO and A-OKVQA datasets. Accuracy and F1 scores are reported for all settings.}
    \label{tab:A:minicpmPope}
    \setlength{\tabcolsep}{5pt}
    \begin{tabular}{lcccc}
        \toprule
        \multirow{2}{*}{Method}
        & \multicolumn{2}{c}{MSCOCO}
        & \multicolumn{2}{c}{A-OKVQA} \\
        \cmidrule(lr){2-3}
        \cmidrule(lr){4-5}
        & Acc. & F1
        & Acc. & F1 \\
        \midrule
        MiniCPM-o 4.5
        & 90.4 & 89.5
        & 92.2 & 91.7 \\
        + LIME (ours)
        & \textbf{92.2} & \textbf{91.9}
        & \textbf{93.85} & \textbf{93.6} \\
        \bottomrule
    \end{tabular}
\end{table}

\begin{table}[t]
    \caption{Evaluation results on HallusionBench for LLaVA-1.5 and Qwen2.5-VL-7B-Instruct. Best and second-best results are highlighted in \textbf{bold} and \underline{underlined}, respectively.}
    \label{tab:hallusionBench}
    \centering
    \begin{tabular}{lcccc}
    \toprule
    \textbf{Methods} & \textbf{fACC $\uparrow$} & \textbf{qACC $\uparrow$} & \textbf{easyA $\uparrow$} &  \textbf{hardA $\uparrow$} \\
    \hline
    LLaVA-1.5 & \underline{17.9} & 8.13 & 36 & 36.7 \\
    + OPERA & 16.2 & 5.49 & \underline{37.6} & 35.4 \\
    + VCD & 13.9 & 11.4 & 33 & 34.7 \\
    + ICD & 13.9 & 8.35 & 36.9 & 33.5 \\
    + MemVR & \underline{17.9} & 9.01 & 36.9 & \textbf{37.7} \\
    + V-ITI & \underline{17.9} & \textbf{10.27} & 36.6 & \underline{37} \\
    + LIME (ours) & \textbf{19.24} & \underline{9.85} & \textbf{37.62} & \textbf{37.7} \\
    \hline
    Qwen2.5-VL-7B-Instruct & 19.29 & 16.4 & 46.11 & 37.9 \\
    + LIME (ours) & \textbf{21.06} & \textbf{18.61} & \textbf{50.37} & \textbf{39.1} \\
    \bottomrule
    \end{tabular}
\end{table}

\begin{figure}[t]
  \centering
  \includegraphics[width=\linewidth]{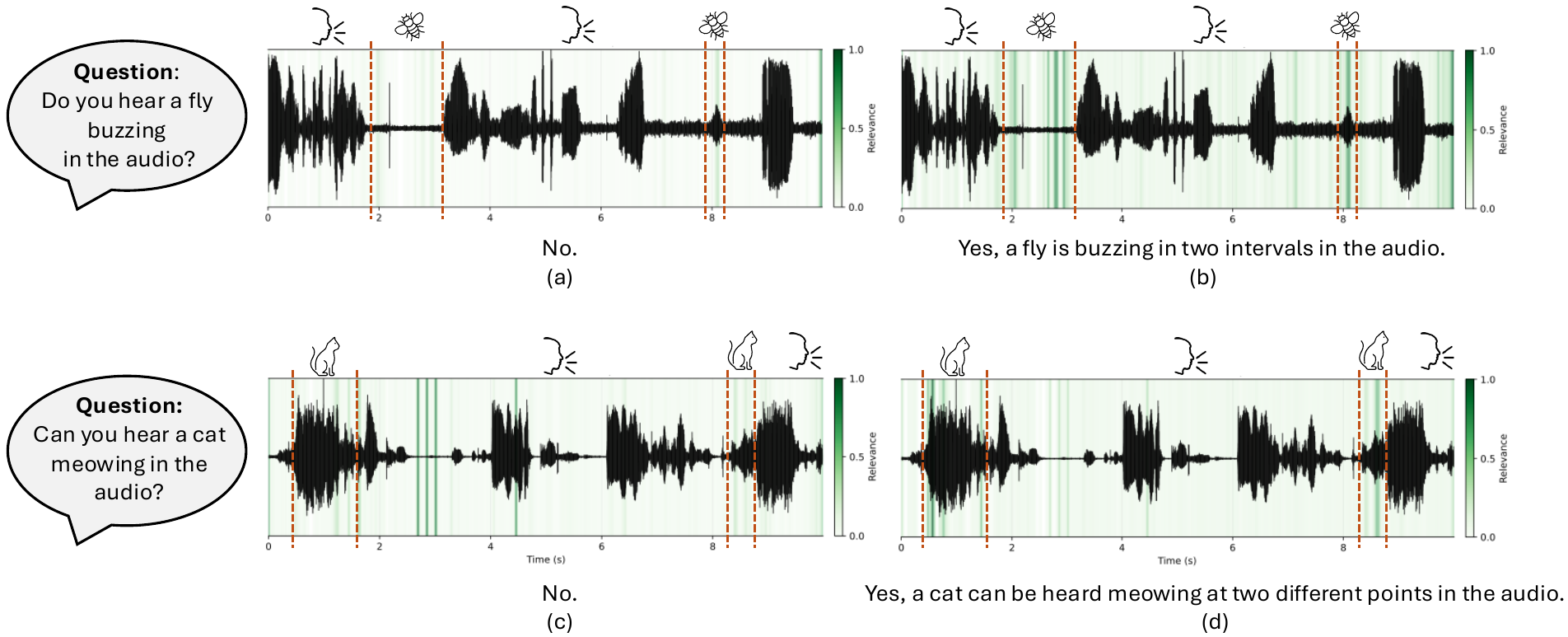}
  \caption{Qualitative examples of multimodal hallucination in the audio domain and its mitigation with LIME. The audio waveform is overlaid with relevance scores (green), and dashed vertical lines indicate ground-truth event regions. Under standard decoding (a,c) relevance is diffuse and poorly aligned with relevant audio segments, leading to incorrect predictions. With LIME (b,d), relevance concentrates on the correct temporal regions, resulting in grounded and accurate predictions.}
  \label{fig:audioExamples}
\end{figure}

\begin{figure}[t]
  \centering
  \includegraphics[width=\linewidth]{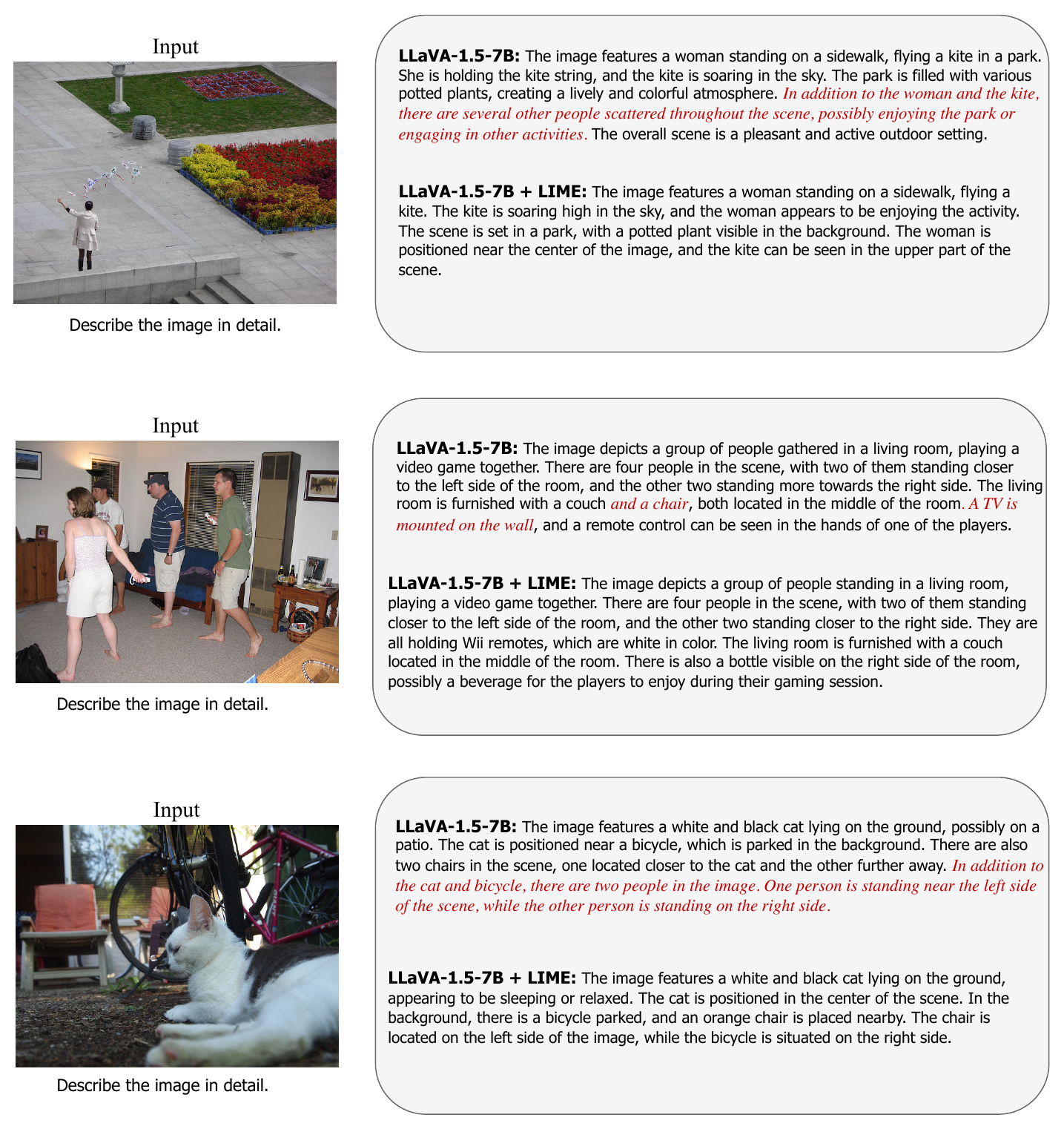}
  \caption{Qualitative examples of hallucination reduction with LIME on LlaVA-1.5-7B. Hallucinated predictions are indicated in \emph{italics} and highlighted in red.}
  \label{fig:llavaExamples}
\end{figure}

\begin{figure}[t]
  \centering
  \includegraphics[width=\linewidth]{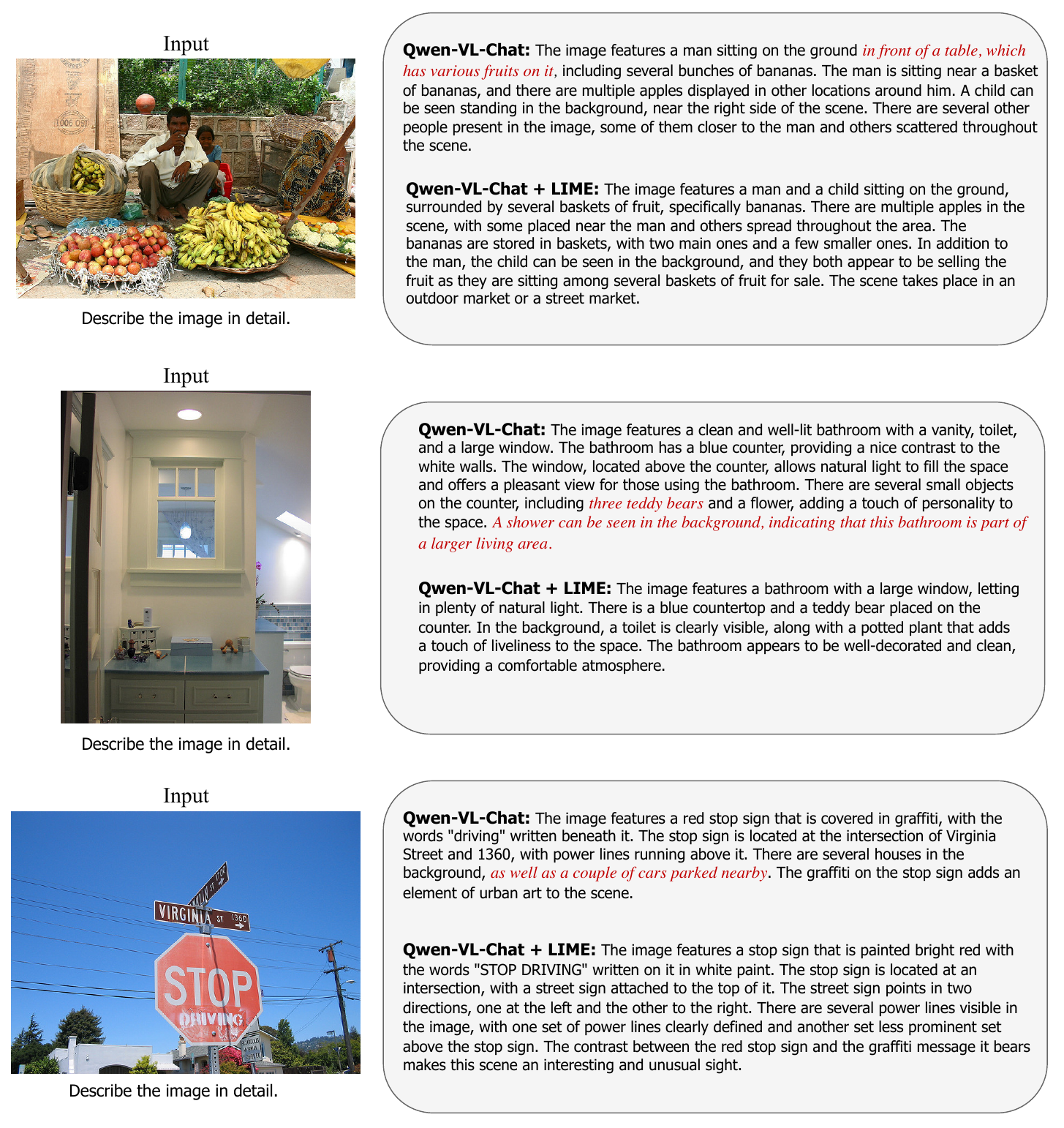}
  \caption{Qualitative examples of hallucination reduction with LIME on Qwen-VL-Chat. Hallucinated predictions are indicated in \emph{italics} and highlighted in red.}
  \label{fig:qwenvlExamples}
\end{figure}

\begin{figure}[t]
  \centering
  \includegraphics[width=\linewidth]{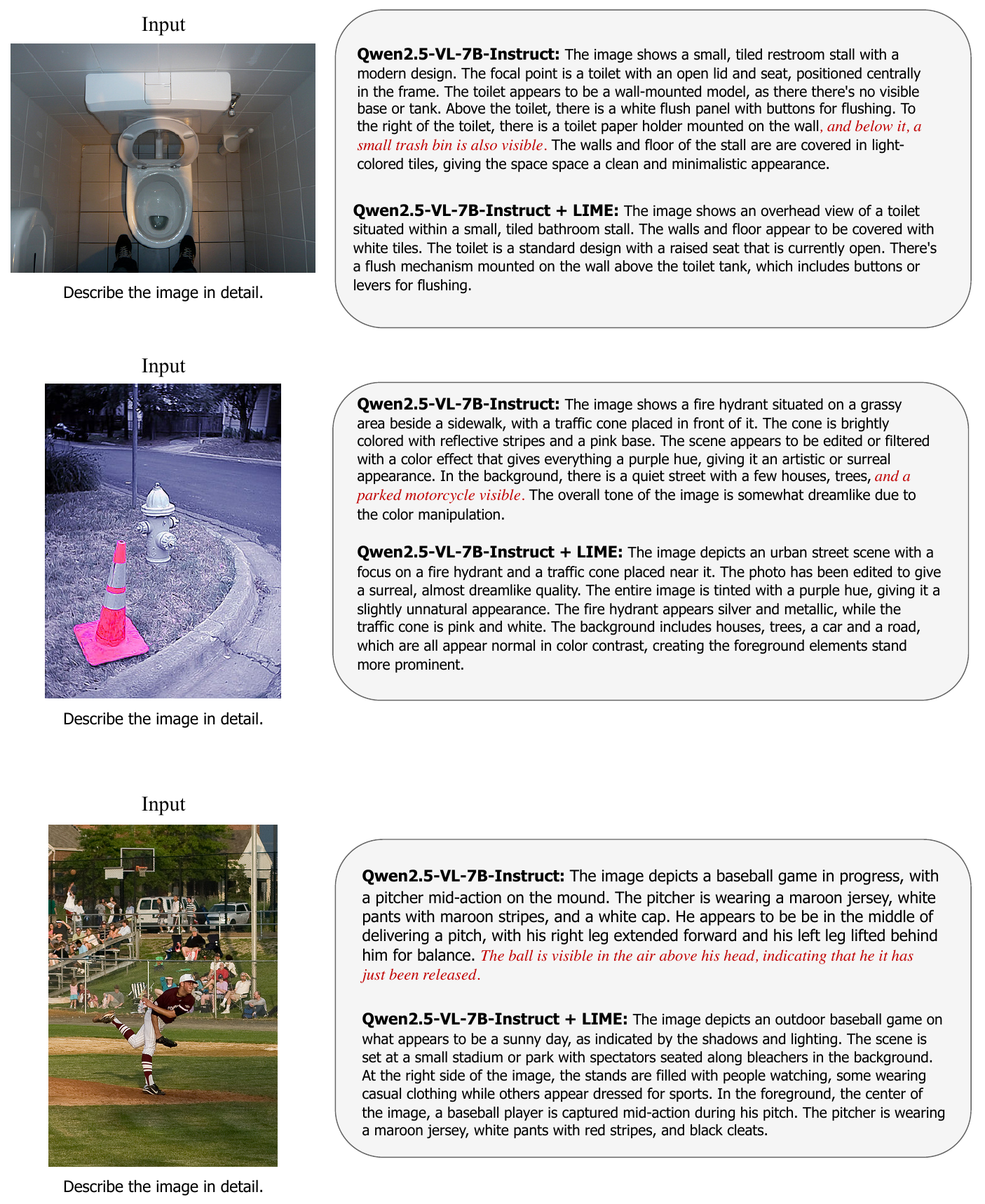}
    \caption{Qualitative examples of hallucination reduction with LIME on Qwen2.5-VL-7B-Instruct. Hallucinated predictions are indicated in \emph{italics} and highlighted in red.}
    \label{fig:qwen2vlExamples}
\end{figure}

\clearpage

\section{Analysis of the Relevance Objective}
\label{app:analysisRelevanceObjective}

The relevance objective in Eq.~\eqref{eq:relevanceLoss} is defined over the aggregate relevance assigned to perceptual and textual tokens, without supervision over which perceptual tokens should gain relevance and without access to prediction correctness. As noted in Section~\ref{sec:inferenceTimeOptimization}, this is a deliberate trade-off for remaining training- and external-model-free: LIME requires no auxiliary verifier, localization module, or labeled data, and can be applied to an MLLM directly at inference time.

This design raises two questions. First, whether the relevance shifts induced by LIME are actually related to changes in hallucination behavior. Appendix~\ref{app:interventional} addresses this through per-sample analysis and controlled interventions on the target and direction of the relevance objective. Second, given that the objective treats all perceptual tokens symmetrically, why does the resulting relevance become concentrated on query-relevant regions? Appendix~\ref{app:mechanism} provides a mechanistic account of how this redistribution is constrained.

\subsection{Interventional Analysis of the Relevance-Hallucination Link}
\label{app:interventional}

Section~\ref{sec:modalityUtilizationAnalysis} shows that LIME increases perceptual utilization and concentrates relevance on query-relevant regions. We next examine whether these relevance shifts are linked to prediction behavior rather than merely accompanying improved performance. We perform two complementary analyses. First, we test whether samples whose predictions are corrected exhibit larger relevance shifts than those that remain hallucinated. Second, we vary which tokens the relevance objective targets and whether their relevance is increased or decreased, while keeping the remaining optimization settings fixed.

\paragraph{Per-sample relevance shift and prediction outcome.}
We use the same 100 POPE samples as in Section~\ref{sec:modalityUtilizationAnalysis} and compute, for each sample, the change $\delta$ in perceptual utilization $\Phi_{\mathcal{P}}/(\Phi_{\mathcal{P}} +
\Phi_{\mathcal{T}})$ and in relevance localization $\sum_{i \in \mathcal{G}} \Phi_i /
\Phi_{\mathcal{P}}$ between vanilla decoding and LIME. Samples are then
partitioned by prediction outcome under the two decoding schemes. Results are reported in Table~\ref{tab:per-sample}, where we give the mean and
standard deviation within each group. Corrected predictions exhibit substantially larger relevance shifts than those
that remain hallucinated: $+0.121$ versus $+0.031$ in perceptual utilization and
$+0.147$ versus $+0.026$ in relevance localization. Samples that were already correct
show intermediate shifts, consistent with the intervention reinforcing grounding
that the model had partially established. The magnitude of the shift therefore
tracks the outcome of individual predictions rather than reflecting only a
co-movement of two group averages. We do not interpret the $\text{correct}
\rightarrow \text{hallucinated}$ group, which contains two samples, we report it
to show that LIME rarely degrades predictions that were already correct.

\begin{table}[t]
\caption{Per-sample relevance shift under LIME relative to vanilla decoding,
grouped by prediction outcome (LLaVA-1.5-7B, POPE). Mean $\pm$ SD are reported.}
\label{tab:per-sample}
\centering
\begin{tabular}{lccc}
\toprule
\textbf{Outcome group} & $\textbf{n}$ & $\delta$ \textbf{Perceptual Utilization} & $\delta$ \textbf{Relevance Localization} \\
\midrule
correct $\rightarrow$ hallucinated      & 2  & $+0.019 \pm 0.033$ & $-0.008 \pm 0.041$ \\
hallucinated $\rightarrow$ hallucinated & 11 & $+0.031 \pm 0.045$ & $+0.026 \pm 0.058$ \\
correct $\rightarrow$ correct           & 64 & $+0.058 \pm 0.041$ & $+0.071 \pm 0.049$ \\
hallucinated $\rightarrow$ correct      & 23 & $\mathbf{+0.121 \pm 0.038}$ & $\mathbf{+0.147 \pm 0.052}$ \\
\bottomrule
\end{tabular}
\end{table}

\paragraph{Sweeping the target and direction of the relevance objective.}
The relevance objective in Eq.~\eqref{eq:relevanceLoss} is defined by a positive set ($\mathcal{P}$) and by a negative sign ($-$). We
vary both while holding everything else fixed: the KL regularizer and its
weight, the layer scope, the optimization budget, the perturbation family, the
learning rate, and the temperature are identical across all variants, so the
configurations differ only in which tokens the objective pushes relevance
toward, and in which direction. We evaluate five variants:
\begin{itemize}
  \item \textbf{LIME}: the method as proposed,
    \[
        \mathcal{L}_{rel}(\Delta) = 
        -\frac{1}{|\mathcal{P}|} \sum_{i \in \mathcal{P}}
          \log \frac{\exp(\Phi_{i,\Delta}/\tau)}
        {\sum_{k \in \mathcal{P} \cup \mathcal{T}} \exp(\Phi_{k,\Delta}/\tau)}
    \]
    
  \item \textbf{Reverse-LIME}: the LIME objective with the sign flipped,
    \[
        \bar{\mathcal{L}}_{rel}(\Delta) = 
        - \mathcal{L}_{rel}(\Delta)
    \]

  \item \textbf{Random}: the same objective with the outer sum taken over a
    random subset $\mathcal{R} \subset \mathcal{P} \cup \mathcal{T}$ with
    $|\mathcal{R}| = |\mathcal{P}|$, resampled at each decoding step,
    \[
        \mathcal{L}^{rand}_{rel}(\Delta) = 
        -\frac{1}{|\mathcal{R}|} \sum_{i \in \mathcal{R}}
        \log \frac{\exp(\Phi_{i,\Delta}/\tau)}
        {\sum_{k \in \mathcal{P} \cup \mathcal{T}} \exp(\Phi_{k,\Delta}/\tau)}
    \]

  \item \textbf{LIME-Sup}: the supervised variant introduced in Eq.~\eqref{eq:supervisedrelevanceLoss}, in which the outer sum is taken only over the ground-truth query-relevant region $\mathcal{G}$,
    \[
        \mathcal{L}^{sup}_{rel}(\Delta) =
        -\frac{1}{|\mathcal{G}|}
        \sum_{i \in \mathcal{P}}
        1_{\{ i \in \mathcal{G}\}} 
        \log \frac{\exp(\Phi_{i,\Delta}/\tau)}
        {\sum_{k \in \mathcal{P} \cup \mathcal{T}}
        \exp(\Phi_{k,\Delta}/\tau)}.
    \]

  \item \textbf{Reverse-LIME-Sup}: LIME-Sup with the sign
    flipped,
    \[
        \bar{\mathcal{L}}^{sup}_{rel}(\Delta) =
        - \mathcal{L}^{sup}_{rel}(\Delta)
    \]
\end{itemize}
As in Section~\ref{sec:modalityUtilizationAnalysis}, $\mathcal{G}$ is available for this analysis only, which allows us to use it as an intervention target here; as stated in Section
\ref{sec:inferenceTimeOptimization}, LIME itself does not assume this prior knowledge. Since the softmax in Eq.~\eqref{eq:relevanceLoss} normalizes over $\mathcal{P} \cup \mathcal{T}$, the resulting relevance shares sum to one, so increasing the share assigned to $\mathcal{P}$ necessarily decreases the share assigned to $\mathcal{T}$. Amplifying $\mathcal{P}$ and suppressing $\mathcal{T}$ are therefore near-equivalent interventions, and we report one variant per pair. We evaluate on the positive split of POPE-MSCOCO with LLaVA-1.5-7B, since $\mathcal{G}$ is defined only for objects that are present, using 5 optimization steps and $\Delta$ on all layers.

Table~\ref{tab:direction-sweep} gives the ordering: Reverse-LIME-Sup $<$
Reverse-LIME $<$ Baseline $\approx$ Random $<$ LIME $<$ LIME-Sup.
Random perturbs with comparable magnitude under an identical objective form yet
performs at baseline ($83.35$ vs.\ $83.24$), so the identity of the amplified tokens, not the presence of an inference-time perturbation, produces the gain. The ordering is signed and monotone about the baseline: amplifying perceptual relevance helps and suppressing it degrades accuracy below the unmodified model. Finally, the effect is asymmetric in $\mathcal{G}$: suppressing relevance on the query-relevant region alone is more damaging ($-3.42$) than suppressing it across all perceptual tokens ($-0.59$), and LIME-Sup improves on LIME ($+1.68$), indicating that the relevance that determines the prediction resides on the query-relevant tokens, rather than being spread across perceptual tokens in general. As in Section~\ref{sec:modalityUtilizationAnalysis}, LIME-Sup bounds what the label-free objective could achieve with perfect knowledge of the region, and the account in Appendix~\ref{app:mechanism} is consistent with our objective recovering much of that gain without access to $\mathcal{G}$. Together, these results show that the reduction in hallucination follows from which tokens gain relevance and which lose it, under otherwise identical optimization.

\begin{table}[h]
\caption{Sweep over the target and direction of the relevance objective
(LLaVA-1.5-7B, positive split of POPE-MSCOCO). All arms share the KL
regularizer, layer scope, optimization budget, and remaining hyperparameters,
and differ only in the positive set and the sign of the objective.}
\label{tab:direction-sweep}
\centering
\begin{tabular}{llcccc}
\toprule
\textbf{Method} & \textbf{Objective target} & \textbf{Direction} & \textbf{Acc (\%)} $\uparrow$ & $\delta$ \textbf{Acc} & \textbf{F1} $\uparrow$ \\
\midrule
Baseline          & ---                     & ---      & 83.24 &  0.00 & 83.12 \\
Random            & random, $|\mathcal{P}|$ & amplify  & 83.35 & +0.11 & 83.20 \\
LIME              & $\mathcal{P}$           & amplify  & 90.44 & +7.20 & 89.88 \\
Reverse-LIME      & $\mathcal{P}$           & suppress & 82.65 & -0.59 & 82.14 \\
Reverse-LIME-Sup  & $\mathcal{G}$           & suppress & 79.82 & -3.42 & 79.45 \\
LIME-Sup          & $\mathcal{G}$           & amplify  & 92.12 & +8.88 & 91.74 \\
\bottomrule
\end{tabular}
\end{table}

\subsection{Mechanism Behind Relevance Localization}
\label{app:mechanism}

Having established that the relevance shift is associated with prediction behavior, we next examine why LIME concentrates relevance on query-relevant perceptual regions despite receiving no spatial or temporal supervision. Section~\ref{sec:modalityUtilizationAnalysis} shows that LIME recovers much of the relevance-localization improvement achieved by LIME-Sup, which explicitly targets the annotated region. This behavior is not directly implied by Eq.~\eqref{eq:relevanceLoss}, since the objective treats all perceptual tokens symmetrically and does not identify a spatial or temporal target.

We develop the explanation in three steps. First, we show that the optimization primarily redistributes relevance and characterize the uniform distribution favored by $\mathcal{L}_{rel}$. Second, we derive the local geometry imposed by the KL constraint. Finally, we show how the frozen model filters the objective's broadly distributed direction, concentrating the realized change along existing perceptual pathways.

\paragraph{Relevance is redistributed, not created.}
LRP conservation (Eq.~\eqref{eq:6}) ties the total input relevance to the logit being explained, $\Phi_{\mathcal{P}}+\Phi_{\mathcal{T}}\approx z_t^*$, while the KL term (Eq.~\eqref{eq:klLoss}) discourages changes to the next-token distribution. Together, these constraints limit changes in the total relevance, so the optimization acts primarily by redistributing relevance among tokens: increasing $\Phi_i$ for some tokens generally requires decreasing it for others. Perceptual utilization measures how this relevance is divided between $\mathcal{P}$ and $\mathcal{T}$, whereas relevance localization measures how the perceptual relevance is distributed within $\mathcal{P}$, specifically how much falls inside the query-relevant subset $\mathcal{G}$.

\paragraph{The objective favors a uniform spread.}
To understand the intrinsic preference of $\mathcal{L}_{rel}$, we first analyze it directly as a function of the relevance scores $\Phi$, abstracting away from how these scores are produced by the model. Let $\vect{s} = \mathrm{softmax}(\Phi/ \tau)$ denote the temperature-scaled relevance distribution, and let
\[
    u_i = \begin{cases}
    \frac{1}{|\mathcal{P}|} & i \in \mathcal{P}, \\
    0 & i \in \mathcal{T},
    \end{cases}
\]
denote the distribution that places equal mass on all perceptual tokens and none on textual tokens. Differentiating Eq.~\eqref{eq:relevanceLoss} with respect to $\Phi$ gives
\begin{equation}
    \vect{g}_{\Phi} = -\nabla_{\Phi}\mathcal{L}_{rel} \;=\; \tfrac{1}{\tau}\,(\vect{u} - \vect{s}),
    \qquad\text{so that}\qquad
    (g_{\Phi})_{i} = \begin{cases}
        \frac{1}{\tau}\Big(\frac{1}{| \mathcal{P}| } - s_i \Big) & i \in \mathcal{P}, \\
        -\frac{1}{\tau}s_i & i \in \mathcal{T}.
        \end{cases}
    \label{eq:relGradient}
\end{equation}
The sign and magnitude of each entry in $\vect{g}_{\Phi}$ indicate how the loss favors changing the relevance of the corresponding token. For textual tokens, $(g_{\Phi})_{i} < 0$, so the loss favors decreasing their relevance. For perceptual tokens, relevance is favored to increase when the current share $s_i$ is below $\frac{1}{|\mathcal{P}|}$, and decreased when it is above this value. Thus, within $\mathcal{P}$, the objective favors equalizing the relevance shares rather than concentrating them on a particular subset.

\begin{remark}
    If the relevance scores were optimized directly under $\mathcal{L}_{rel}$, a stationary point would satisfy $\vect{s} = \vect{u}$, corresponding to equal relevance share across perceptual tokens and zero share on textual tokens. At this point, the fraction of perceptual relevance inside $\mathcal{G}$ would be $\frac{|\mathcal{G}|}{|\mathcal{P}|}$. Therefore, considered on its own, $\mathcal{L}_{rel}$ favors a uniform distribution over $\mathcal{P}$ and does not directly encourage localization to the query-relevant subset $\mathcal{G}$. This mirrors a known pattern in contrastive learning, where the objective alone does not determine what is learned and the architecture and optimizer form part of the account \citep{saunshi2022understanding}.
\end{remark}

\paragraph{The role of the KL term.}
In LIME, the relevance is not optimized directly. Instead, the optimization variables are the key-value perturbations $\Delta$ (Section~\ref{sec:inferenceTimeOptimization}), viewed here as a single vector containing the perturbations across layers. Consequently, the desired change in relevance, represented by $\vect{g}_{\Phi}$, cannot be applied directly. It can only be induced by changing $\Delta$, and propagating its effect through the frozen model, while the KL term penalizes the resulting change in the next-token distribution. We characterize this penalty locally around $\Delta=0$, which is the natural expansion point because LIME reinitializes the perturbations to zero at every decoding step (Section~\ref{sec:inferenceTimeOptimization}).

\begin{lemma}
    \label{lem:fisher}
    Around $\Delta = 0$
    \begin{equation}
        \mathcal{L}_{KL}(\Delta) = \tfrac{1}{2}\Delta^{\top} \mathbf{F} \Delta + O(\|\Delta\|^3),
    \end{equation}
     where $\mathbf{F} \in \mathbb{R}^{D \times D}$ is the Fisher information of the next-token distribution with respect to $\Delta$, evaluated at $\Delta = 0$, and $D$ is the total number of scalar elements in the key-value perturbation vector $\Delta$.
     \begin{equation}
        \mathbf{F} = 
        \mathbb{E}_{y \sim p_\theta}[\nabla_{\Delta}\log p_{\theta,\Delta}\,\nabla_{\Delta}\log
        p_{\theta,\Delta}^{\top}]|_{\Delta = 0}.         
     \end{equation}
\end{lemma}
\begin{proof}
    We expand around $\Delta = 0$, where $p_{\theta,\Delta} = p_{\theta}$. At this point, $\mathcal{L}_{KL}(0)=0$. Its gradient also vanishes:
    \[
        \nabla_{\Delta}\mathcal{L}_{KL}|_{\Delta = 0} =
        \nabla_{\Delta}\sum_y p_{\theta,\Delta}(y) = \nabla_{\Delta} 1 = 0.
    \]
    Differentiating once more and using
    \[
        \nabla_{\Delta} p_{\theta,\Delta} = p_{\theta,\Delta}\nabla_{\Delta}\log p_{\theta,\Delta},
    \]
    the terms involving
    \[
        \sum_{y}\nabla^2_{\Delta}p_{\theta, \Delta}(y) =
        \nabla^2_{\Delta}1 = 0    
    \]
    vanish. The remaining term is
    \[ 
        \mathbb{E}_{y \sim p_{\theta}}[\nabla_{\Delta}\log p_{\theta,\Delta}\,\nabla_{\Delta}\log
        p_{\theta,\Delta}^{\top}]|_{\Delta = 0} = 
        \mathbf{F}.
    \]
    The result follows from the second-order Taylor expansion of $\mathcal{L}_{KL}$ around $\Delta = 0$
\end{proof}
Since both the value and the gradient vanish at $\Delta = 0$, the KL term penalizes the perturbation quadratically in the Fisher metric rather than linearly. The Fisher matrix assigns a higher cost to perturbation directions that more strongly change the next-token distribution.

\paragraph{One optimization step.}
Since relevance depends on the KV perturbation through the frozen model, we write it as $\Phi(\Delta)$. Let $\mathbf{J} = \left. \frac{\partial\Phi(\Delta)}{\partial\Delta} \right|_{\Delta=0} \in\mathbb{R}^{N\times D}$ be its Jacobian at $\Delta=0$, where $N=|\mathcal{P}|+|\mathcal{T}|$ is the number of input tokens. Locally, $\Phi(\Delta)\approx\Phi(0)+\mathbf{J}\Delta$. Using $\vect{g}_{\Phi}=-\nabla_\Phi\mathcal{L}_{rel}$, the corresponding first-order approximation of the relevance loss is
\[
    \mathcal{L}_{rel}(\Phi(\Delta))
    \approx
    \mathcal{L}_{rel}(\Phi(0))
    -
    \vect{g}_{\Phi}^{\top}\mathbf{J}\Delta.
\]
Combining this approximation with Lemma~\ref{lem:fisher} and dropping terms independent of $\Delta$ gives
\[
    \min_{\Delta}
    \left\{
        -\vect{g}_{\Phi}^{\top}\mathbf{J}\Delta
        +
        \tfrac{\lambda}{2}\Delta^{\top}\mathbf{F}\Delta
    \right\}.
\]
On the support of $\mathbf{F}$, its minimum-norm solution is
\[
    \Delta^\star
    =
    \tfrac{1}{\lambda}
    \mathbf{F}^{+}\mathbf{J}^{\top}\vect{g}_{\Phi},
\]
where $\mathbf{F}^{+}$ denotes the pseudoinverse of $\mathbf{F}$. Therefore, the induced change in relevance is
\begin{equation}
    \delta\Phi
    =
    \Phi(\Delta^\star)-\Phi(0)
    \approx
    \vect{J}\Delta^{\star} =
    \tfrac{1}{\lambda}
    \underbrace{
        \mathbf{J}\mathbf{F}^{+}\mathbf{J}^{\top}
    }_{\mathbf{S}}
    \vect{g}_{\Phi}
    =
    \tfrac{1}{\lambda}\mathbf{S}\vect{g}_{\Phi}.
    \label{eq:relStep}
\end{equation}
Thus, the realized relevance change is obtained by filtering the direction favored by the relevance loss, $\vect{g}_{\Phi}$, through the model-dependent matrix $\mathbf{S}$.

\paragraph{Why the update concentrates.}
Eq.~\eqref{eq:relStep} shows that the realized relevance change is not $\vect{g}_{\Phi}$ itself, but $\mathbf{S}\vect{g}_{\Phi}$. Although $\vect{g}_{\Phi}$ distributes the desired increase broadly across perceptual tokens, $\mathbf{S}$ restricts this change according to the sensitivity structure of the frozen model. In particular
\[
    S_{ij}
    =
    \mathbf{J}_{i,:}\mathbf{F}^{+}\mathbf{J}_{j,:}^{\top},
\]
where $\mathbf{J}_{i,:}$ describes how the relevance of token $i$ responds to changes in $\Delta$. Thus, tokens tend to undergo similar relevance changes when they are sensitive to the same perturbation directions allowed by the KL constraint. The broad direction favored by the relevance loss is therefore concentrated along relevance patterns to which the frozen model is already sensitive. The results in Section~\ref{sec:modalityUtilizationAnalysis} provide empirical evidence that the directions favored by the frozen model align with the query-relevant subset $\mathcal{G}$, as the realized relevance increase concentrates within $\mathcal{G}$ despite the objective having no access to this subset.

\section{Efficiency Analysis and Limitations}
\label{app:efficiencyAndlimitations}

Section~\ref{sec:ablationStudies} examines LIME's main design choices, including the roles of key and value updates and the effect of the KL-regularization weight $\lambda$. Here, we extend this analysis to computational efficiency by varying the optimization budget and intervention scope, and by measuring latency and memory overhead. We conclude by discussing the practical limitations of the method.

\begin{figure}[t]
  \centering
  \includegraphics[width=\linewidth]{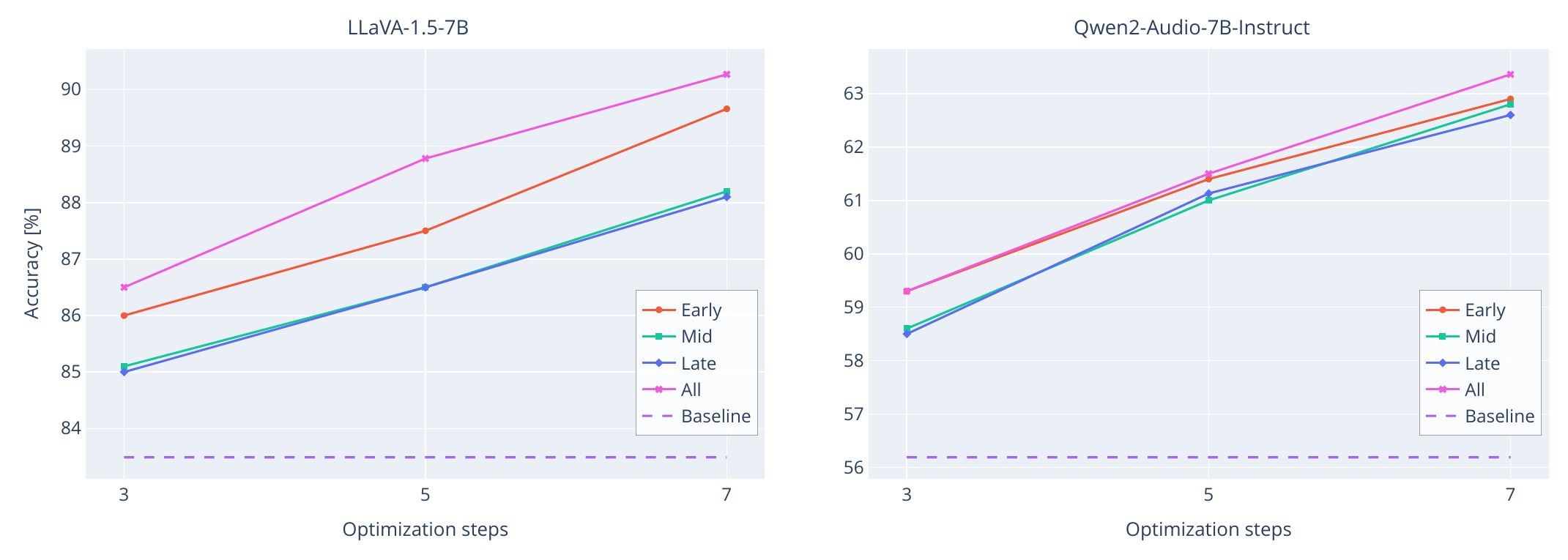}
  \caption{Effect of optimization steps when applying $\Delta$ to early (red), middle (green), late (blue), or all (pink) transformer layers.}
  \label{fig:9}
\end{figure}

\begin{table}[h]
    \caption{Computational overhead comparison between vanilla decoding, representative baselines, and LIME across optimization budgets.}
    \label{tab:A:efficiency}
    \centering
    \begin{tabular}{lcccc}
    \toprule
    \textbf{Methods} & \textbf{Tokens/sec} $\uparrow$ & \textbf{Slowdown} $\downarrow$ & \textbf{Peak Memory (GB)} $\downarrow$ & \textbf{Accuracy (\%)} $\uparrow$\\
    \hline
    LLaVA-1.5-7B & 3.02 & 1.0$\times$ & 26.24 & 83.49\\
    + OPERA & 0.82 & 3.66$\times$ & 39.09 & 87.53\\
    + MemVR & 2.9 & 1.04$\times$ & 26.51 & 88.5\\
    + LIME (5 steps) & 0.45 & 6.73$\times$ & 34.75 & 88.78\\
    + LIME (7 steps) & 0.32 & 9.43$\times$ & 34.75 & 90.27\\
    \hline
    Qwen2-Audio-7B & 2.76 & 1.0$\times$ & 30.84 & 56.19\\
    + AAD & 1.38 & 2.0$\times$ & 32.7 & 59.5\\
    + LIME (5 steps) & 0.42 & 6.57$\times$ & 33.84 & 61.5\\
    + LIME (7 steps) & 0.3 & 9.2$\times$ & 33.84 & 63.36\\
    \bottomrule
    \end{tabular}
\end{table}

\subsection{Efficiency Analysis}
\label{app:efficiency}

\paragraph{Layer scope and optimization budget.}
We investigate whether the $\Delta$ updates must be applied throughout the entire LLM backbone. We consider four intervention scopes: early layers (0-8), middle layers (12-20), late layers (24-32), and all transformer layers. For each scope, we perform three, five, or seven optimization steps. The results are shown in Figure~\ref{fig:9}. Increasing the optimization budget consistently improves accuracy across both models and all intervention scopes. Applying the updates to all layers achieves the strongest performance at every budget. Nevertheless, three optimization steps already provide substantial improvements over the base models, and interventions restricted to selected layer ranges remain effective. For LLaVA-1.5-7B, early-layer updates consistently outperform middle- and late-layer updates, whereas the differences among the restricted scopes are smaller for Qwen2-Audio-7B-Instruct. Overall, these results show that the optimization budget and layer scope provide controllable lower-cost operating points, although the highest accuracy is obtained by applying $\Delta$ throughout the network.

\paragraph{Computational overhead.}
We additionally report the computational overhead introduced by LIME relative to vanilla
decoding and to several representative decoding methods. Table~\ref{tab:A:efficiency} summarizes
throughput, relative slowdown, and peak GPU memory usage measured under the same decoding
configuration used in the main experiments. For LIME, we evaluate five and seven optimization steps with $\Delta$ applied to all transformer layers, corresponding to the strongest intervention scope in the preceding analysis. LIME’s overhead increases with the optimization budget: for LLaVA-1.5-7B, slowdown increases from $6.73\times$ at five steps to $9.43\times$ at seven step, with a similar trend for Qwen2-Audio-7B. In contrast, peak memory remains essentially unchanged across optimization budgets, indicating that the additional cost is primarily in latency rather than memory. The five-step configuration provides a lower-cost operating point, reaching $88.78\%$ accuracy on LLaVA-1.5-7B compared with $90.27\%$ at seven steps, and $61.5\%$ on Qwen2-Audio-7B compared with $63.36\%$. Unlike the representative baselines, which operate with a fixed inference configuration, LIME exposes an explicit optimization budget that allows computational cost to be adjusted: increasing the number of optimization steps improves performance at the cost of higher latency, while fewer steps provide a lower-cost operating point. Together with the layer-scope results above, this makes the efficiency–performance trade-off explicitly controllable.

\subsection{Limitations}
\label{app:limitations}

LIME has several practical limitations. First, although it requires no training, its per-step relevance-guided optimization introduces non-negligible inference-time latency. As shown in Appendix~\ref{app:efficiency}, this cost can be reduced by decreasing the optimization budget or restricting the intervention scope, but at the expense of some performance. Second, the method requires the selection of inference-time hyperparameters, including the KL-regularization weight and optimization budget, whose optimal settings may vary across models and tasks. These factors may limit the applicability of LIME in latency-sensitive or resource-constrained settings.

\end{document}